\documentclass[conference]{IEEEtran}
\IEEEoverridecommandlockouts
\usepackage[hidelinks]{hyperref}
\usepackage{listings, multicol}
\usepackage{algorithm, algorithmicx, algpseudocode}
\usepackage{graphicx}
\usepackage{textcomp}
\usepackage{xcolor}
\usepackage{tabularx,multirow,longtable,booktabs}
\usepackage{balance}
\usepackage{subcaption}
\usepackage{xspace}
\usepackage{makecell}
\usepackage{cite}
\usepackage{amsmath,amssymb,amsfonts}
\usepackage[OT1]{fontenc} 

\newcommand{\sysname}{GLISP\xspace}
\newcommand{\sysfullname}{GLISP(\textbf{G}raph \textbf{L}earning driven by \textbf{I}nherent \textbf{S}tructural \textbf{P}roperties)}
\newcommand{\dne}{DistributedNE\xspace}
\newcommand{\ogbp}{OGBN-Products\xspace}
\newcommand{\ogbpa}{OGBN-Paper\xspace}
\newcommand{\wiki}{WikiKG90Mv2\xspace}
\newcommand{\twitter}{Twitter-2010\xspace}
\renewcommand{\figurename}{Fig.}

\def\BibTeX{{\rm B\kern-.05em{\sc i\kern-.025em b}\kern-.08em
    T\kern-.1667em\lower.7ex\hbox{E}\kern-.125emX}}

\begin{document}
\sloppy
\title{\sysname: A Scalable GNN Learning System by Exploiting Inherent Structural Properties of Graphs}

\author{\IEEEauthorblockN{1\textsuperscript{st} Zhongshu Zhu}
\IEEEauthorblockA{\textit{Ant Group} \\
Hangzhou, China \\
zhongshu.zzs@antgroup.com}
\and
\IEEEauthorblockN{2\textsuperscript{nd} Bin Jing}
\IEEEauthorblockA{\textit{Ant Group} \\
Hangzhou, China \\
jingbin.jb@antgroup.com}
\and
\IEEEauthorblockN{3\textsuperscript{rd} Xiaopei Wan}
\IEEEauthorblockA{\textit{Ant Group} \\
Hangzhou, China \\
ranghou.wxp@antgroup.com}
\and
\IEEEauthorblockN{4\textsuperscript{th} Zhizhen Liu}
\IEEEauthorblockA{\textit{Ant Group} \\
Hangzhou, China \\
zhizhen.lzz@antgroup.com}
\and
\IEEEauthorblockN{5\textsuperscript{th} Lei Liang}
\IEEEauthorblockA{\textit{Ant Group} \\
Hangzhou, China \\
leywar.liang@antgroup.com}
\and
\IEEEauthorblockN{6\textsuperscript{th} Jun zhou}
\IEEEauthorblockA{\textit{Ant Group} \\
Hangzhou, China \\
jun.zhoujun@antfin.com}
}

\maketitle

\begin{abstract}
As a powerful tool for modeling graph data, Graph Neural Networks (GNNs) have received increasing attention in both academia and industry. Nevertheless, it is notoriously difficult to deploy GNNs on industrial scale graphs, due to their huge data size and complex topological structures. In this paper, we propose \sysname, a sampling based GNN learning system for industrial scale graphs. By exploiting the inherent structural properties of graphs, such as power law distribution and data locality, \sysname addresses the scalability and performance issues that arise at different stages of the graph learning process. \sysname consists of three core components: graph partitioner, graph sampling service and graph inference engine. The graph partitioner adopts the proposed vertex-cut graph partitioning algorithm AdaDNE to produce balanced partitioning for power law graphs, which is essential for sampling based GNN systems. The graph sampling service employs a load balancing design that allows the one hop sampling request of high degree vertices to be handled by multiple servers. In conjunction with the memory efficient data structure, the efficiency and scalability are effectively improved. The graph inference engine splits the $K$-layer GNN into $K$ slices and caches the vertex embeddings produced by each slice in the data locality aware hybrid caching system for reuse, thus completely eliminating redundant computation caused by the data dependency of graph. Extensive experiments show that \sysname achieves up to $6.53\times$ and $70.77\times$ speedups over existing GNN systems for training and inference tasks, respectively, and can scale to the graph with over 10 billion vertices and 40 billion edges with limited resources. 
\end{abstract}


\section{Introduction}
\label{sec:intro}

As a non-Euclidean data structure, graphs can capture complex relationships between entities and are widely used in modeling real world data, such as knowledge graphs\cite{wangKnowledgeGraphEmbedding2017}, web graphs\cite{donatoLargeScaleProperties2004a}, and social networks\cite{yingGraphConvolutionalNeural2018}. In recent years, Graph Neural Networks (GNNs)\cite{kipfSemiSupervisedClassificationGraph2016, hamiltonInductiveRepresentationLearning2018, liuGeniePathGraphNeural2018, shumanEmergingFieldSignal2013, velickovicGraphAttentionNetworks2018, zhouGraphNeuralNetworks2019} have achieved great success in various graph learning tasks, such as vertex classification and link prediction. However, the scale of real world graphs is typically large and growing rapidly, which poses a serious challenge to the application of GNNs. For example, the Taobao dataset contains millions of vertices and tens of millions of edges\cite{Jizhe2018Billion}, while the User-to-Item graph of Pinterest comprises billions of vertices and edges\cite{yingGraphConvolutionalNeural2018}. For these huge graphs, full-graph training of GNN is intractable due to the limitation of CPU/GPU memory. An alternative is sampling based mini-batch training\cite{zengGraphSAINTGraphSampling2020, chiangClusterGCNEfficientAlgorithm2019, qiuGCCGraphContrastive2020} where batches of small subgraphs are sampled from the original graph as model inputs. A variety of GNN frameworks have introduced such mini-batch training methods, including DGL\cite{wangDeepGraphLibrary2020}, PyTorch Geometric\cite{feyFastGraphRepresentation2019}, etc. However, subject to the hardware resources of a single machine, it is still challenging for these frameworks to handle very large scale graph data.

\begin{figure}[!htbp]
  \centering
  \includegraphics[width=0.7\columnwidth]{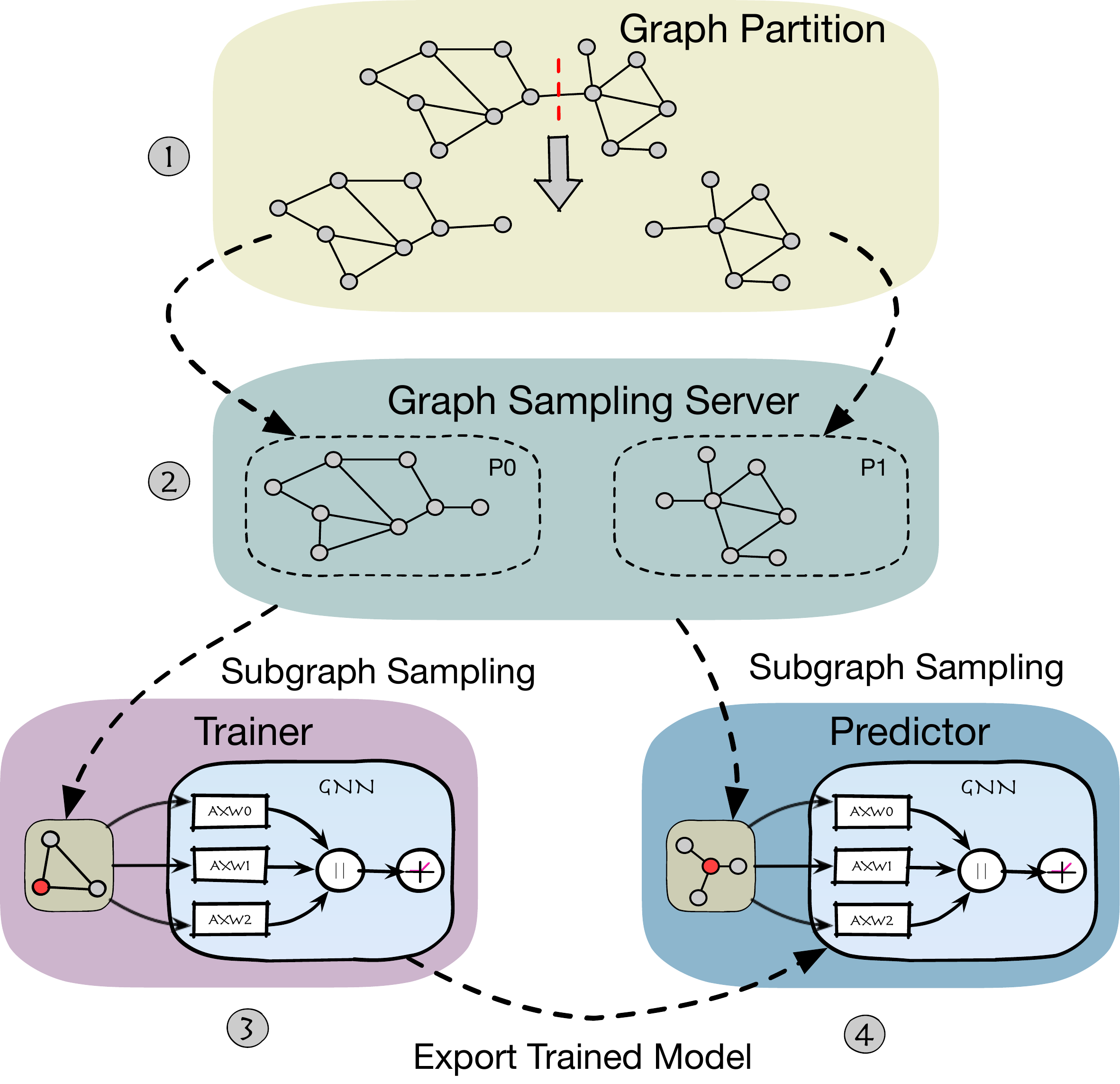}
  \caption{\label{fig:online-framework} Schematic diagram of the distributed graph learning framework. A complete workflow consists of four steps: graph partitioning, launching graph sampling service, training and inference.}
\end{figure}

Distributed frameworks such as DistDGL\cite{wangDeepGraphLibrary2020, zhengDistDGLDistributedGraph2020, zhengDistributedHybridCPU2022}, GraphLearn\cite{zhuAliGraphComprehensiveGraph2019}, Euler\cite{EulerGithub2022}, BGL\cite{liuBGLGPUEfficientGNN2021} and ByteGNN\cite{zhengByteGNNEfficientGraph} have been developed to tackle the challenge. \autoref{fig:online-framework} is the workflow of the distributed frameworks. The graph is first partitioned into multiple parts and a distributed graph sampling server is launched in conjunction with the trainer/predictor to generate subgraphs. After training, one can run inference on the whole graph to obtain the vertex embeddings for downstream tasks.

Although the scale of graph data in the real world is growing rapidly, the size of the GNN model has not increased correspondingly due to the ubiquitous \textit{over smoothing} problem\cite{liDeeperGCNAllYou2020, liDeeperInsightsGraph2018, zhouGraphNeuralNetworks2019, chenMeasuringRelievingOversmoothing2019}. Extensive studies have revealed that deeply stacking GNN layers can lead to a substantial decrease in model performance\cite{liDeeperInsightsGraph2018, oonoGraphNeuralNetworks2021, caiNoteOverSmoothingGraph2020}. Due to the huge gap between data size and model size, the bottleneck of graph learning tasks is mainly on the data side, such as graph partitioning and subgraph sampling. Therefore, we give an analysis of the limitations of the existing frameworks from the perspective of graph data:

\textbf{Graph partitioning is not optimized for sampling based graph learning task on power law graphs.} As shown in \autoref{fig:online-framework}, the big graph is first partitioned to multiple parts, each can be fitted into a single machine. There are two mainstream graph partition algorithms, i.e.,  \textit{edge-cut}\cite{karypisFastHighQuality, karypisParallelAlgorithmMultilevel1998, slotaPartitioningTrillionedgeGraphs2016} and \textit{vertex-cut}\cite{hanaiDistributedEdgePartitioning2019, kongClusteringbasedPartitioningLarge2022} partitioning. The edge-cut partitioning divides the vertices into disjoint parts while the vertex-cut partitioning divides the edges disjointly. Existing distributed graph learning frameworks all adopt the edge-cut partitioning(DistDGL, GraphLearn , Euler, BGL, ByteGNN, DSP\cite{cai2023dsp}, etc.) where vertices and their one hop neighbors are located in the same partition via redundant cut edges, so as to avoid the inter-partition communication in the one hop neighbor sampling algorithm. On the other hand, most real world graphs follow power law distributions\cite{crucittiErrorAttackTolerance2004, donatoLargeScaleProperties2004, gonzalezPowerGraphDistributedGraphParallel2012} and contain several high degree vertices, i.e., hotspots. Theoretically and experimentally, it has been proved that edge-cut partitioning on power law graphs can lead to severe data skewness and redundancy\cite{ gonzalezPowerGraphDistributedGraphParallel2012, hanaiDistributedEdgePartitioning2019, kongClusteringbasedPartitioningLarge2022}.  Also, unlike the graph processing systems and full graph GNN systems\cite{wanPipeGCNEfficientFullGraph2021, ImprovingAccuracyScalability}, whose performance is bounded by cross-partition communication, the sampling-based GNN systems follow the Client-Server architecture, and the data flows from the graph server to GNN trainer/predictor, not between servers, so \textit{no cross-partition communication is involved and workload balance dominates the performance}. Unfortunately, existing partition algorithms do not fully consider the distinctive data access pattern.

\textbf{Neighbor sampling architectures are load imbalanced and memory inefficient.} The skewed degree distribution of power law graphs can also lead to the load imbalance of distributed sampling service. Systems such as DistDGL, BGL, and ByteGNN strive to attain load balancing by distributing the training vertices evenly across servers. However, \textit{simply balancing input seeds is inadequate}. Specifically, existing frameworks tailored for edge-cut partitioned graphs assign one-hop neighbor sampling requests for each vertex to a single server. Servers containing hotspots are accessed more frequently in $K$-hop neighbor sampling, resulting in higher overhead. Hence, even with perfectly balanced seeds, load imbalances will still occur as $K$ increases. Vertex-cut partitioning can effectively balance the workload by allocate one hop neighbors of hotspots to multiple partitions, at the cost of re-implementing the one-hop sampling algorithm in a distributed manner. To our knowledge, no sampling framework based on vertex-cut partitioning has been proposed. In addition, the intricacy of graph data poses a challenge to the data structure design. For example, many real world graphs are heterogeneous, and the recent proposed heterogeneous sampling strategies require indexing by edge type\cite{wangHeterogeneousGraphAttention2019, fuMAGNNMetapathAggregated2020}. DistDGL and GraphLearn represent the heterogeneous graph by multiple homogeneous graphs, one for each edge type, resulting in high memory footprint and non-contiguous memory layout. Euler stores the edge type ID for each edge and builds the edge type index for each vertex’s neighbors separately, which also brings additional memory consumption.

\textbf{Inference on the whole giant graph is intractable.} For industrial scale graphs, the inference phase may take more time than the training phase for several reasons. First, the training set only accounts for a small fraction of the whole dataset. For instance, the OGB-Paper\cite{huOpenGraphBenchmark2021} dataset contains 111M vertices, while the training set size is only 1.2M. Second, in real-world applications such as recommendation and fraud detection, the graph structure and vertex/edge features are constantly updated, and the trained model periodically performs inference on the updated graph, resulting in a computational workload that far exceeds the training task. Existing frameworks do not fully consider the characteristics of the full graph inference task, making it difficult to scale to giant graphs. For example, contrary to the training phase, the model parameters are fixed in the inference phase, which introduces considerable redundant computation due to the overlap between the $K$-hop neighbors of multiple vertices.

Based on these observations, we propose \sysfullname, a scalable GNN learning system for real world power law graphs consists of three core components: \textbf{graph partitioner}, \textbf{graph sampling service} and \textbf{graph inference engine}. \sysname exploits the inherent structural properties of graphs, such as power law distribution and data locality, to address the aforementioned issues of existing GNN systems.

We highlight the following key contributions:

\begin{itemize}

\item We propose AdaDNE, a vertex-cut based graph partition algorithm for graph learning tasks on industrial scale power law graphs. AdaDNE adaptively adjusts the partitioning speed of each worker by imposing soft constraints on the number of vertices and edges within each partition, resulting in a significant improvement in vertex and edge balance while maintaining low redundancy.

\item We propose a load balanced neighbor sampling architecture. The $K$ hop sampling are formalized as $K$ iterative Gather-Apply based one hop sampling operations, where each one hop sampling of hotspots can be handled collaboratively by multiple servers. The data structure for the vertex-cut partitioned graphs adopts a contiguous memory layout and substantially reduces the memory footprint by replacing some fields with $O(1)$ or $O(\log N)$ time complexity queries.

\item We propose a redundant computation free graph inference engine, where the computation of $K$ layer GNNs is divided into $K$ slices and the intermediate vertex embeddings are cached in the hybrid static and dynamic caching system for scalability. The data locality is fully exploited by our graph reorder algorithm PDS to accelerate the large scale embedding retrieval.

\end{itemize}

\section{Preliminaries}
\label{sec:background}

\subsection{Sampling based GNN Training}
\label{subsec:background-gnn}

GNNs are a family of neural networks designed for data described by graphs. Given the graph structure and vertex/edge features as input, the GNNs can learn a vector representation for each vertex in the graph, which can be used as the input to the downstream tasks. Most of GNNs follow the message passing paradigm, where vertex representations are updated by gathering the message of their neighbors iteratively. Formally speaking, given a graph $G = \{V, E\}$, the GNN computes the vector representation of vertex $v$ at layer $i+1$ by the following equation:

\begin{equation}
\label{eq:gnn}
\mathbf{h}_v^{i+1}=\sigma^i(\mathbf{h}_v^{i}, Agg_{u\in\mathcal{N}(v)}(\mathbf{h}_v^{i}, \mathbf{h}_u^i, \mathbf{e}_{u,v}))
\end{equation}
where $\mathbf{h}_v^{i}$ is the vector representation of vertex $v$ at the $i$th layer, when $i=0$, it denotes the input vertex feature. And $\mathcal{N}(v)$ is the set of neighbor vertices of $v$, $\mathbf{e}_{u,v}$ is the feature of edge $(u,v)$, $\sigma^i$ and $Agg$ are the learnable feature update and aggregation function of the $i$th layer of GNN, respectively.

Since $K$-layer GNN is equivalent to performing $K$ message passing over the graph, the final vector representation $\bf{h}_v^K$ of vertex $v$ depends on its $K$ hop neighbors. As $K$ increases, the size of the neighboring vertices grows exponentially, making the computation intractable. 

\begin{figure}[!htbp]
  \centering
  \includegraphics[width=0.8\columnwidth]{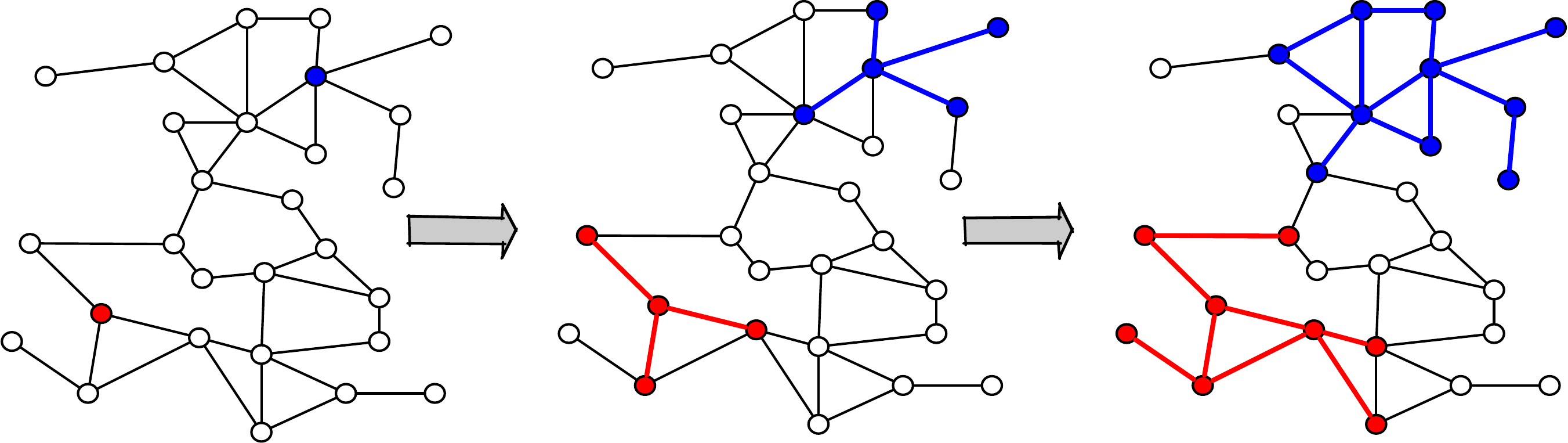}
  \caption{\label{fig:khop-neighbor-sampling} Schematic diagram of $K$ hop neighbor sampling, where $K=2$ and 2 seed vertices are selected.}
\end{figure}

One way to alleviate the exponential explosion of $K$ hop neighbors in GNN computation is to use mini-batch neighbor sampling, i.e., sampling a small-sized subgraph from the original graph as the model input. \autoref{fig:khop-neighbor-sampling} gives an example neighborhood sampling process. First, we randomly select a target vertex $v$ from the dataset $V$, and then sample at most $f_1$ vertices from $v$’s one hop neighbors ${\mathcal{N}(v)}$. Starting from the selected neighbor vertices, one can continue to sample up to $f_2$ neighbors for each vertex. The above sampling process is repeated recursively until hop $K$ is reached. The target vertex $v$ is called seed and the parameters $\mathcal{F}=[f_1, f_2, \dots, f_K]$ are called fanouts. Various neighbor sampling strategies have been proposed, such as weighted neighbor sampling\cite{zengGraphSAINTGraphSampling2020} and meta-path sampling for heterogeneous graphs\cite{fuMAGNNMetapathAggregated2020, yingGraphConvolutionalNeural2018}.

\subsection{Graph Partition} 
\label{subsec:background-graph-partition}

\begin{figure}[!htbp]
  \centering
  \includegraphics[width=0.8\columnwidth]{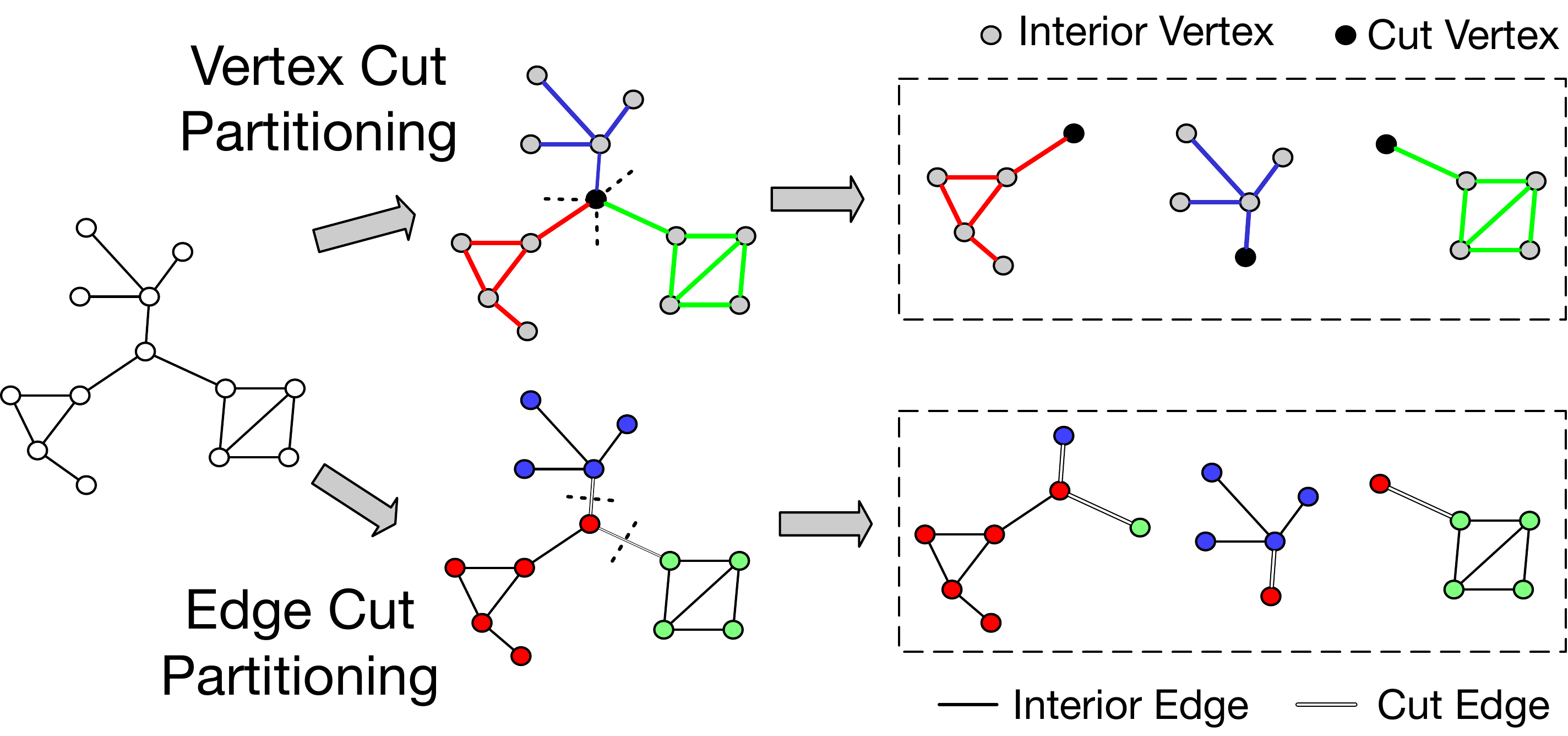}
  \caption{\label{fig:vertex-edge-partition} Schematic diagram of vertex-cut and edge-cut partition.}
\end{figure}

Graph partitioning aims to divide the input graph $G=\{V, E\}$ into $|P|$ similar-sized subgraphs while minimizing the overall vertex/edge redundancy, the $p$th subgraph $G_p$ contains $|V_p|$ vertices and $|E_p|$ edges\cite{karypisFastHighQuality, hanaiDistributedEdgePartitioning2019, karypisParallelAlgorithmMultilevel1998, slotaPartitioningTrillionedgeGraphs2016, fan2020incrementalization}. There are two mainstream graph partition algorithms, vertex-cut partition\cite{hanaiDistributedEdgePartitioning2019, kongClusteringbasedPartitioningLarge2022} and edge-cut partition\cite{karypisFastHighQuality, karypisParallelAlgorithmMultilevel1998, slotaPartitioningTrillionedgeGraphs2016}. As shown in \autoref{fig:vertex-edge-partition}, the vertex-cut partition assigns a partition ID to each edge $e$, resulting in vertices being cut, and these cut vertices (also known as \textit{boundary vertices}) are stored redundantly on multiple parts. In contrast, the edge-cut partition assigns a partition ID to each vertex $v$, resulting in edges being cut as well as redundant storage.

The graph partition problem has been proved to be NP-hard\cite{andreevBalancedGraphPartitioning2004} and existing algorithms are mainly based on heuristic strategies, such as random partition, 1D/2D Hash, METIS\cite{karypisFastHighQuality, karypisParallelAlgorithmMultilevel1998}, PuLP\cite{slotaPuLPScalableMultiobjective2015, slotaPartitioningTrillionedgeGraphs2016}, Distributed Neighbor Expansion(\dne)\cite{hanaiDistributedEdgePartitioning2019}. Partitioning quality is evaluated by means of redundancy and balance. The Replication Factor ($RF$) defined in \eqref{eq:rf} reflects the redundancy, and the Edge Balance ($EB$) in \eqref{eq:eb} and Vertex Balance ($VB$) in \eqref{eq:vb} are used to measure the imbalance of edges and vertices, respectively. A high quality partition algorithm should keep these metrics as close to 1 as possible. High $RF$ means more data redundancy and high memory footprint, whereas high $EB$ and $VB$ indicate data skew and workload imbalance of each 
part.

\begin{equation}
  \label{eq:rf}
  RF = \frac{\sum_{p\in P}|V_p|}{|V|}
\end{equation}

\begin{equation}
  \label{eq:eb}
EB = \frac{\max_{p \in P}(|E_p|)}{\min_{p \in P}(|E_p|)}
\end{equation}

\begin{equation}
  \label{eq:vb}
  VB = \frac{\max_{p \in P}(|V_p|)}{\min_{p \in P}(|V_p|)}
\end{equation}

\subsection{Graph Reorder} 
\label{subsec:background-graph-reorder}

Graph reorder is a class of algorithms that exploit the locality of graph data by seeking a new arrangement for the vertex IDs such that spatially close vertices have close IDs\cite{cuthillReducingBandwidthSparse1969,  blandfordExperimentalAnalysisCompact, dhulipalaCompressingGraphsIndexes2016, SpeculativeParallelReverse, haoSpeedupGraphProcessing2016, araiRabbitOrderJustinTime2016, balajiWhenGraphReordering2018, barikVertexReorderingRealWorld2020}. For example, vertex $v$ and its neighbor $\mathcal{N}(v)$ are expected to be assigned consecutive IDs. Many downstream tasks can benefit from graph reorder. To name a few, one can compress the graph by graph reorder and delta-encoding techniques\cite{boldiWebgraphFrameworkCompression2004, dhulipalaCompressingGraphsIndexes2016}. Graph reorder can also improve the cache hit ratio of neighbor accesses, thereby speeding up many graph computation tasks\cite{balajiWhenGraphReordering2018, barikVertexReorderingRealWorld2020, haoSpeedupGraphProcessing2016}.

Various graph reorder algorithms have been proposed, such as the heavyweight Gorder\cite{haoSpeedupGraphProcessing2016}, Recursive Graph Bisection (RGB)\cite{dhulipalaCompressingGraphsIndexes2016}, Reverse Cuthill-McKee (RCM)\cite{cuthillReducingBandwidthSparse1969}, and the lightweight BFS, Degree Sort(DS), Rabbit\cite{araiRabbitOrderJustinTime2016}, Hub Clustering\cite{balajiWhenGraphReordering2018}, etc.

It is worth noting that graph partition algorithms can also be viewed as a class of graph reorder algorithms, i.e., partitioning-based schemes\cite{barikVertexReorderingRealWorld2020}, since vertices can be assigned to different partitions and ordered by partition ID.

\section{System Design}
\label{sec:system-design}
\subsection{Architecture} 
\label{subsec:system-overview}

\begin{figure}[!htbp]
  \centering
  \includegraphics[width=\columnwidth]{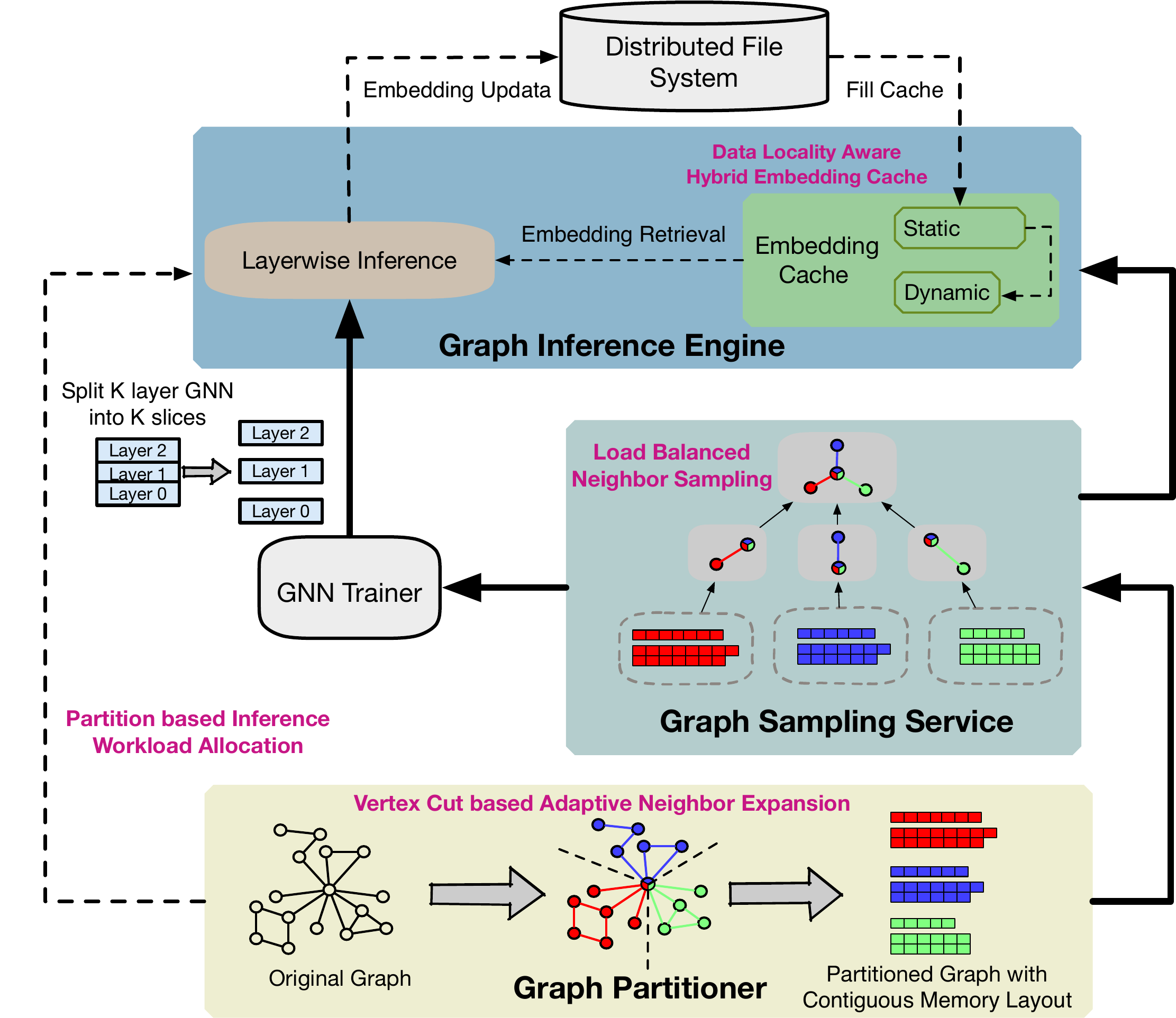}
  \caption{\label{fig:architecture} The system architecture of \sysname.}
\end{figure}

The architecture of \sysname, as shown in \autoref{fig:architecture}, comprises three components: graph partitioner, graph sampling service, and graph inference engine. Instead of implementing a new GNN modeling component, we directly adapted mature GNN libraries such as PyG\cite{feyFastGraphRepresentation2019} and DGL\cite{wangDeepGraphLibrary2020}. As the cornerstone of the whole system, the \textbf{graph partitioner} implements graph partition algorithm optimized for sampling based graph learning task of power law graphs, at the heart of which is \textit{vertex-cut adaptive neighbour expansion}. The \textbf{graph sampling service} provides an contiguous and memory efficient data structure for the vertex-cut partitioned graph, and adopts a \textit{Gather-Apply paradigm based load balancing architecture} to generate input subgraphs. The \textbf{graph inference engine} computes the the vertex embeddings layer by layer to eliminate redundant computation. The intermediate vertex embeddings produced by each GNN layer are stored in the Distributed File System(DFS) for reuse. By exploiting the data locality mined by graph partitioning and graph reordering, the graph inference engine integrates a \textit{two-level hybrid embedding caching system} to speedup large scale embedding retrieval from remote.

\subsection{Graph Partitioner} 
\label{subsec:system-graph-partitioner}

Existing partitioning algorithms prioritize reducing $RF$ to minimize cross-partition communication, which is valid for graph processing systems and full graph GNN systems\cite{wanPipeGCNEfficientFullGraph2021, ImprovingAccuracyScalability}. In sampling based GNN systems, however, $RF$ has \textbf{limited impact on the performance since data flows from server to GNN trainer/predictor, rather than between servers}. Instead, load balancing plays a critical role in system performance\cite{zhengDistDGLDistributedGraph2020, liuBGLGPUEfficientGNN2021, zhengByteGNNEfficientGraph}, so a balanced partitioning is required, not only the vertex balance (as indicated by DGL, BGL and ByteGNN), but also the edge balance, since the workload essentially depends on the number of edges from the graph computing perspective\cite{hanaiDistributedEdgePartitioning2019}.

Following the above analysis, we propose the Adaptive Distributed Neighbor Expansion (AdaDNE) algorithm that employs the \textbf{vertex-cut} partitioning as opposed to the edge-cut to handle real world power law graphs and \textbf{prioritize the balanced partitioning}. AdaDNE introduces several key optimizations on top of \dne\cite{hanaiDistributedEdgePartitioning2019} that simultaneously suppress the edge and vertex imbalances. We first give a brief review of \dne, and then present AdaDNE optimizations.

The core of \dne is neighbor expansion, i.e., starting from a randomly selected seed vertex, iteratively adding its neighbor edges to the partition. In a nutshell, \dne consists of the following steps:

\begin{itemize}

\item Initialize: Generate initial partitions by 2D-Hash partitioning, and randomly select the initial vertex set in each partition. The neighbors of the initial vertex set become the boundary vertex set $B_p$.

\item Neighbor Expansion: Select $\lambda|B_p|$ vertices with the smallest degree $V_p$, where the constant $\lambda$ is the expansion factor used to control the percentage of vertices involved in expansion, i.e., the expansion speed.

\item One Hop Edge Allocation: Allocate the one hop neighbor edges of $V_p$ to current partition, and add the one hop neighbor vertices of $V_p$ to $B_p$.

\item Two Hop Edge Allocation: For the one hop neighbor edge $e(u, w)$ of $B_p$ (i.e., two hop neighbor edge of $V_p$), if the two endpoints of $e(u, v)$ have already been allocated to the same partitions $P_{uv} = \{P_u\} \cap \{P_v\}$,  then allocate $e(u, v)$ to the minimal edge partition $argmin_{p \in P_{uv}} |E_p|$.

\item Check Termination: If the number of edges $|E_p|$ of a partition exceeds the threshold $E_{t}=\tau\frac{|E|}{|P|}$, the partition is terminated, where the constant $\tau$ is the imbalance factor. Otherwise, goto Neighbor Expansion for the next iteration.

\end{itemize}

With the hard constraint introduced by $E_t$, \dne guarantees that $EB$ is close to 1, but $VB$ can be very large due to \textit{the absence of constraint on the number of vertices}. We try to add the missing constraint on $VB$ to AdaDNE. Intuitively, to obtain a balanced partition, the expansion speed of partition $p$ should decrease if the number of vertices/edges of $p$ exceeds the average, and vice versa. However, \dne adopts the same expansion speed for all partitions, which is defined by the constant $\lambda$. 

We modify the expansion speed control policy of \dne to an adaptive approach. In particular, at the $i+1$th iteration of partition $p$, the vertex score $VS_p^i$ and edge score $ES_p^i$ of $p$ are first updated by \eqref{eq:vb-p} and \eqref{eq:eb-p}, respectively. Then the adaptive expansion factor $\lambda_p^{i+1}$ of $p$ can be calculated by \eqref{eq:lambda-p}, where $\alpha$ and $\beta$ are the weight of vertex and edge score, respectively. Since the adaptive expansion factor is now a function of the vertex/edge score and can serve as a soft constraint on the number of vertices/edges, the edge threshold $E_t$ in \dne is then removed. In other words, this is equivalent to setting $\tau=|P|$.

\begin{equation}
  \label{eq:vb-p}
  VS_p^i = \frac{|P||V_p|}{\sum_{n\in P}|V_p|}
\end{equation}

\begin{equation}
  \label{eq:eb-p}
  ES_p^i = \frac{|P||E_p|}{\sum_{n\in P}|E_p|}
\end{equation}

\begin{equation}
  \label{eq:lambda-p}
  \lambda_p^{i+1} = \lambda_p^i*\exp(\alpha(1-VS_p^i)+\beta(1-ES_p^i))
\end{equation}

AdaDNE updates the adaptive expansion factor of each partition by simply synchronizing the number of vertices and edges of all partitions at the beginning of each iteration. This process incurs negligible computational and communication overhead. Moreover, AdaDNE follows the heuristic neighbor expansion strategy of \dne, so the theoretical analysis on the partitioning quality and efficiency of \dne is also valid for AdaDNE and will not be repeated here\cite{hanaiDistributedEdgePartitioning2019}.

\subsection{Graph Sampling Service} 
\label{subsec:system-graph-sampling-service}

One key feature of the vertex-cut partitioning is that it allows the one hop neighbors of a vertex to be distributed across multiple partitions. Therefore, \textbf{the one-hop neighbor sampling task of a single vertex can be done cooperatively by multiple servers}, thus making it possible to balance the server workload. 

\begin{figure}[!htbp]
  \centering
  \includegraphics[width=0.85\columnwidth]{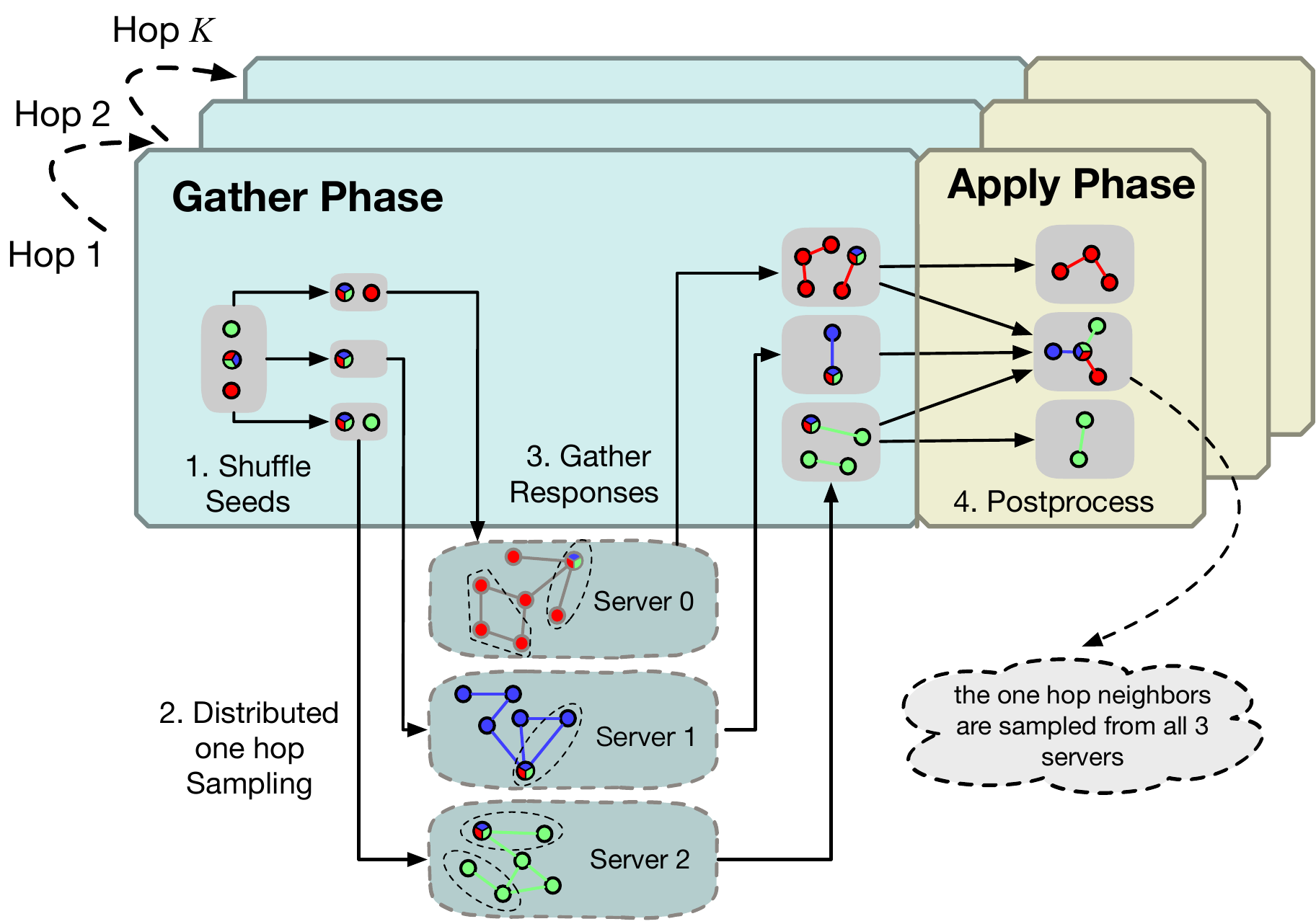}
  \caption{\label{fig:sampling-framework} The Gather-Apply based $K$ hop neighbor sampling algorithm. The key difference from existing frameworks is the one hop sampling requests for the boundary vertices are handled cooperatively by all servers on which it resides to substentially balance the workload, as shown in the second seed vertex (tri-colored marker).}
\end{figure}

Motivated by the Gather-Apply-Scatter (GAS) programming model\cite{gonzalezPowerGraphDistributedGraphParallel2012}, we formalize the distributed $K$ hop uniform/weighted neighbor sampling algorithm as $K$ \textbf{Gather} and \textbf{Apply} operations. \autoref{fig:sampling-framework} shows an illustrative example of the $K$ hop neighbor sampling algorithm. For graph containing $P$ partitions, $P$ servers will be launched, each for one partition. In the \textbf{Gather} phase, the client distributes one hop sampling requests to all servers containing the seed vertices and gathers the responses. Sampling requests are processed independently by each server on the local partitioned graph. In the \textbf{Apply} phase, the client post-processes the partial sampled one hop neighborers from each server to obtain the final result. Typically, a hotspot's neighbors exist on almost all servers, so the distributed one hop neighbor sampling strategy can substantially balance the workload. Algorithm \autoref{algo:neighbor-sampling} gives the pseudo-code of the $K$ hop neighbor sampling, where the uniform and weighted neighbor sampling are accomplished by setting different $GatherOp$ and $ApplyOp$, respectively.

\begin{algorithm}
\caption{$K$ hop Uniform/Weighted Neighbor Sampling.}
\label{algo:neighbor-sampling}
\begin{algorithmic}[1]
\renewcommand{\algorithmicrequire}{\textbf{Input:}}
\renewcommand{\algorithmicensure}{\textbf{Output:}}
\Require Seed vertices $\mathcal{S}$, fanouts $\mathcal{F}$, sampling configurations $\mathcal{C}$
\Ensure Sampled $K$ hop subgraph $G_{\mathcal{S}}$
\If{$C.weighted$}
    \State $GatherOp \leftarrow WeightedGatherOp$
    \State $ApplyOp \leftarrow WeightedApplyOp$

\Else:
    \State $GatherOp \leftarrow UniformGatherOp$
    \State $ApplyOp \leftarrow UniformApplyOp$

\EndIf
\For{$f\in \mathcal{F}$} 
    \State $N_f \leftarrow [\mathbf{Gather}(GatherOp, [\mathcal{S}, C,f, p]),\ \mathbf{for} \  p=1...P]$ 
    \State $G_f\leftarrow \mathbf{Apply}(ApplyOp, [S,C,f,G_S, N_f])$
    \State $\mathcal{S} \leftarrow GetSeedsOfNextHop(G_f)$
    \State $G_\mathcal{S}.append(G_f)$
\EndFor

\State return $G_{\mathcal{S}}$
\end{algorithmic} 
\end{algorithm}

For uniform neighbor sampling, each server can calculate the number of edges to be sampled according to the fanout $f$, the global degree and the local degree of vertex, where the global and local degrees are the vertex degrees in the \textit{original} graph and the \textit{partitioned} graph, respectively. Then the AlgorithmD\cite{vitterEfficientAlgorithmSequential1987b} can be used to perform uniform sampling. Algorithm \autoref{algo:uniform-sampling} is the implementation of $UniformGatherOp$. We omit the $UniformApplyOp$ since it is simply a matter of joining the results together.

\begin{algorithm}
\caption{$UniformGatherOp$ (Server Side)}
\label{algo:uniform-sampling}
\begin{algorithmic}[1]
\renewcommand{\algorithmicrequire}{\textbf{Input:}}
\renewcommand{\algorithmicensure}{\textbf{Output:}}
\renewcommand{\algorithmicprocedure}{\textbf{Function:}}
\Require Seed vertices $\mathcal{S}$, sampling configurations $\mathcal{C}$, fanout $f$, current partition $p$
\Ensure Sampled neighbors of current partition $N_p$

\For{$s \in \mathcal{S}$}
    \State $local\_deg \leftarrow GetLocalDegree(s, p, \mathcal{C}.direction)$
    \State $global\_deg \leftarrow GetGlobalDegree(s, \mathcal{C}.direction)$

    \State $r\leftarrow f*local\_deg/global\_deg$
    \State $n\leftarrow AlgorithmD(s, C.direction, r)$
    \State $N_p.append(n)$
\EndFor
\State return $N_p$
\end{algorithmic} 
\end{algorithm}

The case of weighted neighbor sampling is a bit more complicated, as it requires knowledge of all neighbor weights. The alias method\cite{walkerNewFastMethod1974}, a well-known technique for constant time sampling from arbitrary probability distributions $p_i$, is not suitable for our scenario. Firstly, constructing a static alias table for each vertex's neighbors requires significant memory and lacks flexibility. Secondly, for the vertex-cut partitioning, a complicated distributed alias method is required.

Our weighted neighbor sampling adopts the AlgorithmA-ES\cite{efraimidisWeightedRandomSampling2006} to address the above issues of alias method. Given the candidate set $\{x_1, x_2\dots x_N\}$ and weights $\{w_1, w_2\dots w_N\}$, AlgorithmA-ES samples $n$ items without replacement by following steps. First, generate a uniform random number $u_i$ between $[0,1]$ for each item $x_i$, and calculate the score $s_i = u_i^{\frac{1}{w_i}}$. Then, select the $n$ samples with the highest score.

AlgorithmA-ES reduces weighted sampling to the Top-K problem, where the distributed version can be easily implemented within the Gather-Apply paradigm. Algorithm \autoref{algo:weighte-sampling-gather} and Algorithm \autoref{algo:weighte-sampling-apply} provide the implementation of $WeightedGatherOp$ and $WeightedApplyOp$, respectively, where the core is the distributed AlgorithmA-ES shown in line 3 of Algorithm \autoref{algo:weighte-sampling-gather} and Algorithm \autoref{algo:weighte-sampling-apply} .

\begin{algorithm}
\caption{$WeightedGatherOp$ (Server Side)}
\label{algo:weighte-sampling-gather}
\begin{algorithmic}[1]
\renewcommand{\algorithmicrequire}{\textbf{Input:}}
\renewcommand{\algorithmicensure}{\textbf{Output:}}
\Require Seed vertices $\mathcal{S}$, sampling configurations $\mathcal{C}$, fanout $f$, current partition $p$
\Ensure Sampled neighbors and corresponding scores of current partition $N_p$
\For{ $s \in \mathcal{S}$}
    \State $w \leftarrow GetWeights(s, C.directtion, C.weight\_conf)$
    \State $n,score\leftarrow AlgorithmAES(s, C.direction, f)$ 
    \State $N_p.append([n, score])$
\EndFor
\State return $N_p$

\end{algorithmic} 
\end{algorithm}

\begin{algorithm}
\caption{$WeightedApplyOp$ (Client Side)}
\label{algo:weighte-sampling-apply}
\begin{algorithmic}[1]
\renewcommand{\algorithmicrequire}{\textbf{Input:}}
\renewcommand{\algorithmicensure}{\textbf{Output:}}
\Require Seed vertices $\mathcal{S}$, sampling configurations $\mathcal{C}$, fanout $f$, sampled neighbors $N_f$
\Ensure Sampled one hop subgraphs $G_S$ of $\mathcal{S}$
\For{ $s \in \mathcal{S}$}
    \State $\{n_i, score_i\}\leftarrow CollectNeighbors(N_f, s)$
    \State $n\leftarrow GetScoreTopK(\{n_i, score_i\}, f)$
    \State $G_S.add(n)$
\EndFor
\State return $G_S$

\end{algorithmic} 
\end{algorithm}

\begin{figure}[!htbp]
  \centering
  \includegraphics[width=0.8\columnwidth]{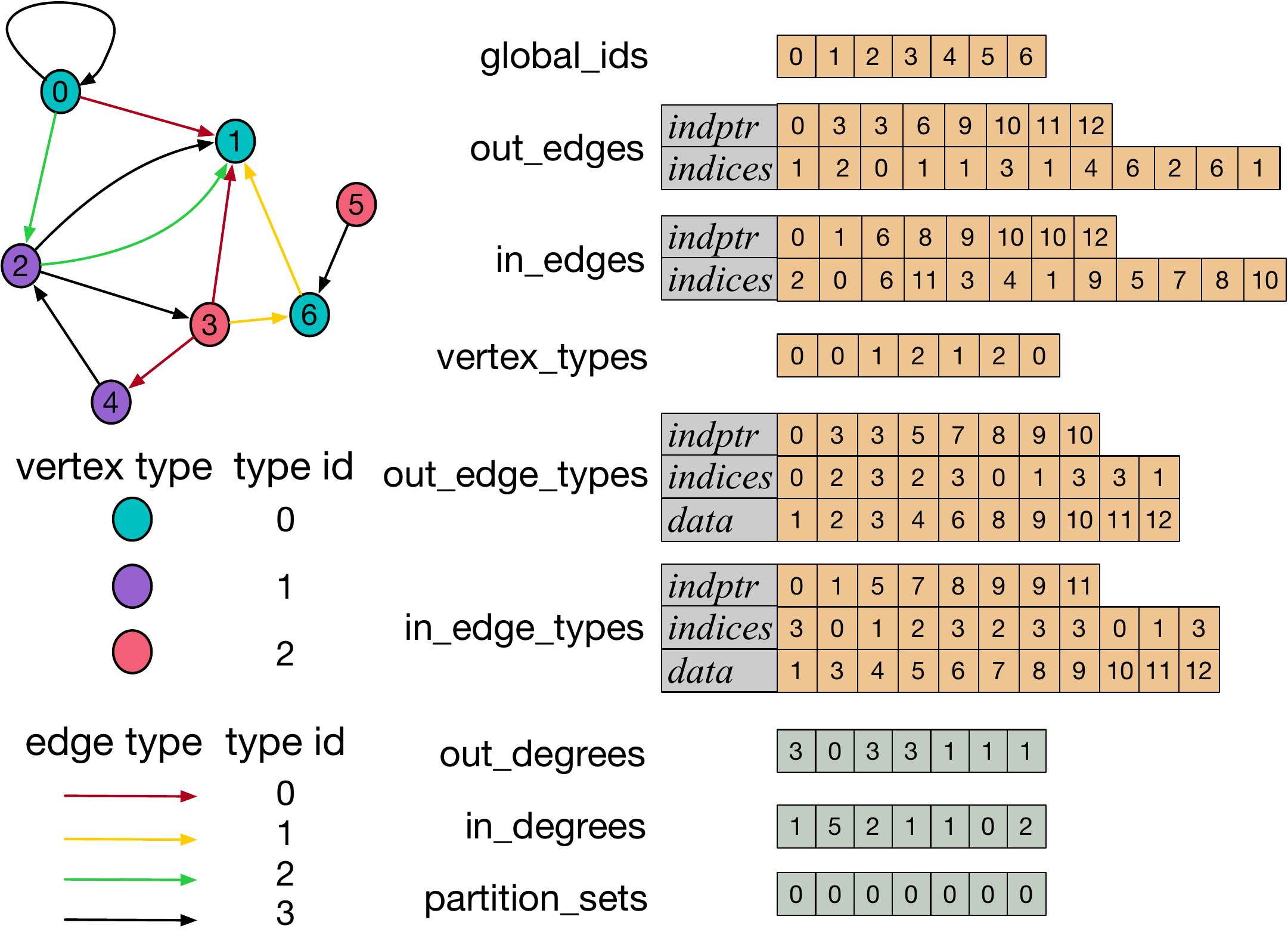}
  \caption{Schematic diagram of the graph data structure. The well-known Compressed Sparse Row (CSR) format consists of $indptr$, $indice$ and an optional $data$ field.\label{fig:data-structure} }
\end{figure}

Another key consideration for the graph sampling service is the graph data structure. A well-designed data structure must address memory consumption and performance issues while meeting the requirements of the distributed sampling algorithm described above. Based on the fact that the server's cpu utilization is considerably lower than memory, we propose a read-only data structure for the the vertex-cut partitioned graph, with the distinctive features of being \textbf{properly sorted} and \textbf{contiguous in memory}. Some necessary fields such as vertex/edge local ID and edge type IDs are replaced by queries with $O(1)$ or $O(\log N)$ time complexity, which effectively reduces the memory overhead. Experiments show that the additional query time is typically only $1\%$ of the total sampling time, which is insignificant. \autoref{fig:data-structure} is a schematic diagram of the data structure for a small heterogeneous multigraph with 7 vertices, 12 edges, 3 vertex types and 4 edge types, with further details provided below.

The graph structure is defined by global\_id, out\_edges and in\_edges, where the global\_id stores the global IDs of all vertices in the partitioned graph in \textit{ascending order}, and the out\_edges and in\_edges are the outgoing and incoming edges, respectively. Since the global IDs of vertices are not consecutive in the partitioned graph, we need to assign consecutive local IDs to vertices to encode the edges and store vertex related attributes such as degree and type. The same concept applies to edges, where edge local ID is used to access edge-related attributes, such as timestamp for dynamic graph. Existing frameworks explicitly assign local IDs to vertices and edges within each partition, resulting in excessive memory consumption. In our data structure, the vertex local ID is implicitly defined by the position index in global\_ids and bidirectional mapping between global and local IDs can be achieved by simple array access and binary search. Similarly, the edge local ID is defined by the position index of out\_edges.  Meanwhile, we replace the edges stored in in\_edge from $(dst\_id, src\_id)$ to $(dst\_id, edge\_id)$, enabling direct access to edge IDs for incoming edges.

The out(in) edge type index of the heterogeneous graph is embedded in out(in)\_edges and out(in)\_edge\_types. Taking the out edges as an example, we first sort the edges in field out\_edge by triple $(src\_id, edge\_type, dst\_id)$ to ensure that the one hop neighbors of each vertex are grouped by edge type. Next, we count the number of edges per type for each vertex, and store the result in out\_edge\_types. Among them, the $indices$ holds the edge type IDs of each vertex, and $data$ stores the number of edges corresponding to each edge type. We have pre-accumulated the $data$, so that the ranges of each edge type in out\_edges can be directly obtained. The $indptr$ stores the offset of each vertex as usual. There is no need to store the individual type ID for each edge as it can be accessed by binary search from the aggregated representation. The type index for in-edges is constructed in the same way.

The remaining 3 fields in \autoref{fig:data-structure} are used to meet the requirements of the distributed neighbor sampling. In particular, out\_degrees and in\_degrees stores the global out-degree and in-degree of vertices, respectively. The partition\_set maintains the partition IDs of each vertex in the format of bit array. 

The data structure is stored and loaded in a simple contiguous binary layout, with the data size and type of each field being maintained in a separate meta file. Since the implementation does not introduce complex containers like HashMap and Vector, there is no additional memory overhead. Furthermore, the one hop neighbors of each vertex are stored consecutively, thus enabling efficient memory access.

\subsection{Graph Inference Engine} 
\label{subsec:system-full-graph-inference}

The graph inference engine aims to efficiently performing GNN inference on graphs with billions or even tens of billions of vertices. Contrary to the training task, the model parameters are no longer updated during inference, thus exhibit the following distinctive features:
\begin{itemize}
  \item output vertex embeddings can be reused
  \item there is no need for randomness in sample ordering, which is required in Stochastic Gradient Descent (SGD)
\end{itemize}

\begin{figure}[!htbp]
  \centering
  \includegraphics[width=0.9\columnwidth]{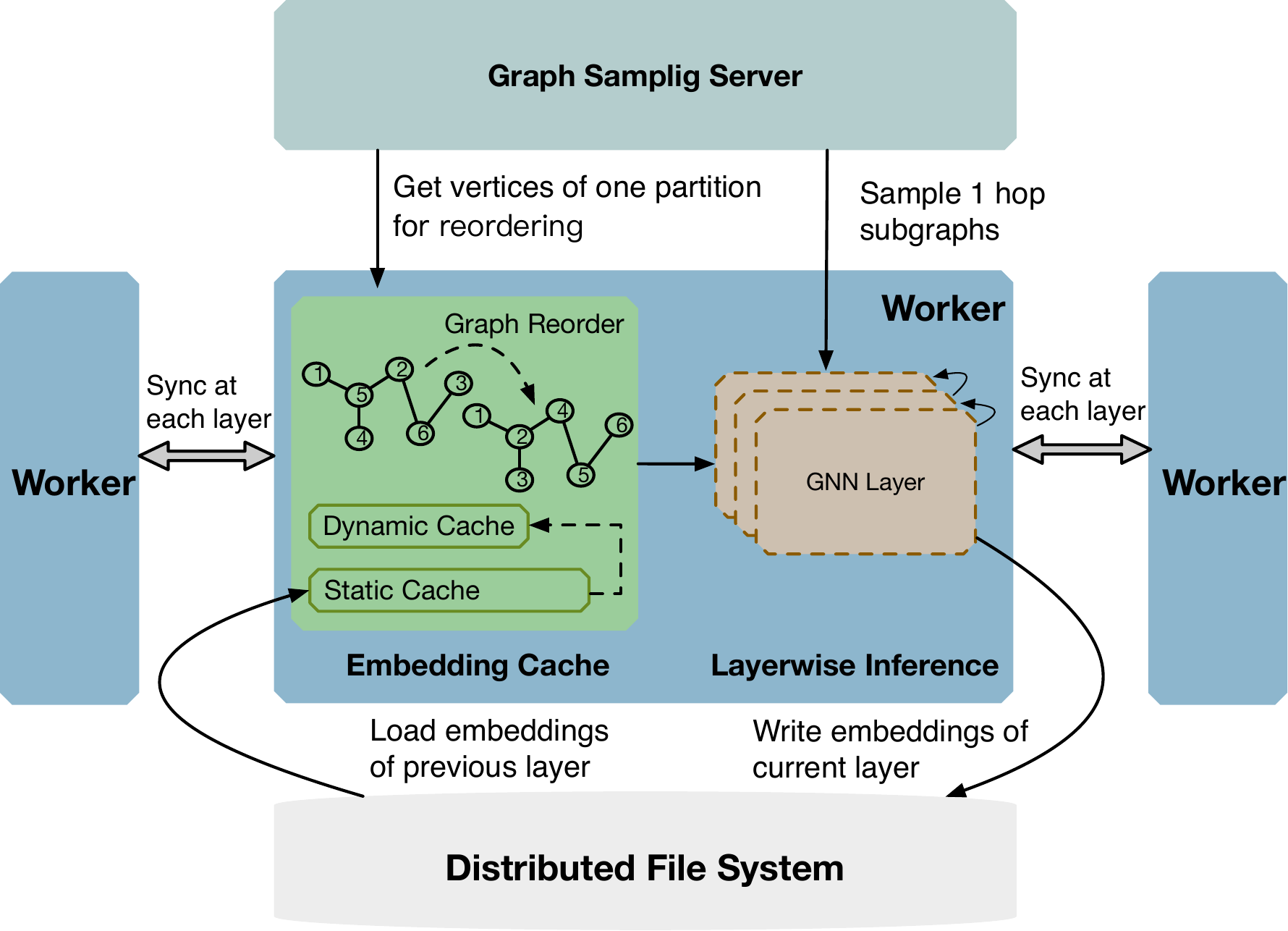}
  \caption{\label{fig:full-graph-infer-framework} The architecture of graph inference engine. The $K$-layer GNN is split into $K$ one-layer GNNs, where vertex embeddings are computed layer by layer to eliminate redundant computations. The intermediate embeddings are cached on DFS for scalability. To speedup embedding reads, the graph inference engine incorporates a data locality aware two-level hybrid caching system, and the graph reordering algorithm PDS is adopted to improve data locality.}
\end{figure}

\autoref{fig:full-graph-infer-framework} shows the architecture of the graph inference engine. The key designs are the \textbf{Layerwise Inference Scheme} and the \textbf{Embedding Caching System}, which are based on the above two features of the inference task, respectively. Specifically, the layerwise inference scheme eliminates the redundant computation by reusing vertex embeddings\cite{yingGraphConvolutionalNeural2018, zhangAGLScalableSystem2020, yinDGIEasyEfficient2023}, while the embedding caching system accelerates the embedding retrieval by partitioned based workload allocation and graph reordering.

Given a $K$ layer GNN model, the naive samplewise inference independently feeds the $K$ hop subgraph of each sample into the model, yielding the target vertex embedding of the $K$th GNN layer. All vertices embeddings of intermediate GNN layers are discarded. If a vertex appears in the $K$ hop neighbors of several target vertices, it's embedding is recalculated several times. In contrast, by storing intermediate results for later use, the layerwise inference scheme completely eliminates redundant computation. Specifically, the $K$-layer GNN is split into $K$ one-layer GNNs, each of which traverses all vertices in the graph to compute the vertex embeddings. Accordingly, the $K$-hop neighbor sampling procedure is split into $K$ 1-hop neighbor sampling. The inputs to the $k$th one-layer GNN are the one hop neighbors of the vertices and the output embeddings of the $k-1$th layer. 

We use the Zarr\cite{ZarrPythonZarr14} format to store the intermediate embeddings, where the entire embedding matrix is chunked into multiple files and compressed with Blosclz clevel 9\cite{BloscMainBlog}. For industrial scale data, these embedding files are stored on DFS and fetching large scale vertex embeddings remotely becomes the system bottleneck. Nevertheless, since there are no restrictions on the randomness of input samples, we are free to exploit the data locality that naturally exists in graphs. Therefore, we design a \textbf{Two-Level Embedding Caching System} consisting of a static disk-based cache and a dynamic memory-based cache to remove the bottleneck.

The \textbf{first level} static disk cache leverages the data locality mined by the graph partitioner. In particular, we first allocate the inference workloads (i.e., all vertices in the graph) according to the partitioned graph, one partition per worker, and then cache the embedding of all vertices in their respective partitions. If $v$ is an interior vertex of partition $i$, then its one hop neighbors $\mathcal{N}(v)$ are all located in partition $i$ and the embedding retrieval of $v$'s one hop subgraph can hit the static cache. For the remaining boundary vertices, we precompute the one hop sampled neighbors and additionally cache the embedding of neighbor vertices that reside in other partitions. In this way, \textit{a $100\%$ cache hit ratio can be guaranteed by simply filling the static cache before inference of each GNN layer}. Benefit from the skewed degree distribution of the power law graph, the majority of vertices in AdaDNE partitioned graph are interior vertices(typically $>70\%$) and the number of additionally cached vertices is acceptable.

By layerwise inference, we replace the repeated vertex embedding computation with the repeated cache reads. For instance, given an edge $e(u, v)$, we read the embeddings of both endpoints $v$ and $w$ twice to update their embeddings, respectively. The obversion inspired us to develop a memory based \textbf{second level} cache on top of the static disk cache. By dynamically caching a fraction of the vertex embeddings with update policy such as LRU or FIFO, the embedding retrieval speed can be further improved.

We need to assign consecutive local IDs to the cached vertices for embedding IO, including the vertices from within the partition and the sampled neighbors of boundary vertices from other partitions. The inference order within each worker is specified by the consecutive local IDs as well. An intuitive idea is to reuse the scheme in the graph sampling service, where local IDs are implicitly defined by the order of global IDs. However this (equivalently) random ordering cannot fully exploit the data locality, so we employ the graph reorder algorithm to generate local IDs of vertices. For industrial scale graph, even a single partition can contain hundreds of millions of vertices. Although heavyweight graph reorder algorithms such as RGB and RCM can yield high-quality results, they incur excessive runtime overhead (typically tens of hours)\cite{balajiWhenGraphReordering2018}.

As stated in \autoref{subsec:background-graph-reorder}, graph partitioning is also a class of graph reordering, so an implicit reordering is actually done by the graph partitioner. It is obviously unwise to reorder the cached vertices from scratch with lightweight algorithms like DS or BFS. We propose a lightweight \textbf{Partition based Degree Sort}(PDS) algorithm. PDS groups vertices by partition ID and applies the DS algorithm within each group, which is functionally equivalent to sorting vertices by the pair $(partition\_id, degree)$. By taking advantage of the data locality mined by the graph partitioner, PDS can produce high-quality arrangement with negligible overehead.

After the reordering, $v$ and $\mathcal{N}(v)$ will have similar IDs. The benefit is twofold:
\begin{itemize}
\item The total number of chunks read during inference is reduced.
\item It is more likely that the same vertex embedding will be accessed repeatedly in a short period of time, which helps to improve the hit ratio of dynamic cache.
\end{itemize}

\section{Evaluation}
\label{sec:evaluation}

In this section, we have conducted a series of experiments to benchmark the performance of \sysname on datasets of different scales, and compared it with the latest open source frameworks DistDGL (V1.1.2)\cite{wangDeepGraphLibrary2020, zhengDistDGLDistributedGraph2020, zhengDistributedHybridCPU2022}, PaGraph(V0.1)\cite{linPaGraphScalingGNN2020} and GraphLearn(V1.1.0)\cite{zhuAliGraphComprehensiveGraph2019}. Since PgGraph is developed on top of the outdated DistDGL V0.4 and lacks support for distributed neighbor sampling, we upgrade the DistDGL version that PaGraph relies on to V1.1.2. All experiments are run on the internal heterogeneous CPU/GPU cluster. We will apply for different numbers of workers for different datasets and experiments. Unless otherwise specified, the default configuration of each worker is  32 vCPU cores * 64GB memory and the GPU worker has a Tesla P100-PCIE-16GB GPU.

\subsection{Datasets}
\label{sec:eval-dataset}

\autoref{table:datatset-stat} is the statistics of the dataset used in our experiments, which contains three publicly available datasets of: \ogbp\cite{huOpenGraphBenchmark2021}, \wiki\cite{huOGBLSCLargeScaleChallenge2021}, \twitter\cite{Twitter2010}, \ogbpa\cite{huOpenGraphBenchmark2021}, and an internal large scale user relationship dataset RelNet.

\begin{table}[!htbp]
  \centering
  \caption{Statistics of datasets.}
  \label{table:datatset-stat}
  \begin{tabular}{>{\centering}p{0.21\columnwidth}|>{\centering}p{0.22\columnwidth}|>{\centering}p{0.22\columnwidth}|>{\centering}p{0.18\columnwidth}}
    \hline
    Dataset&\# Vertices&\# Edges&Average Degree \tabularnewline \hline
    \ogbp&2,449,029&61,859,140&25.2 \tabularnewline \hline
    \wiki&91,230,610&601,062,811&6.6\tabularnewline \hline
    \twitter&41,652,230&1,468,365,181&35.3\tabularnewline \hline
    \ogbpa&111,059,956&1,615,685,872&14.5 \tabularnewline \hline
    RelNet&10,458,278,953&48,959,724,102&4.7\tabularnewline \hline
\end{tabular}
\end{table}

\begin{figure*}[!htbp]
  \centering
  \includegraphics[width=2\columnwidth]{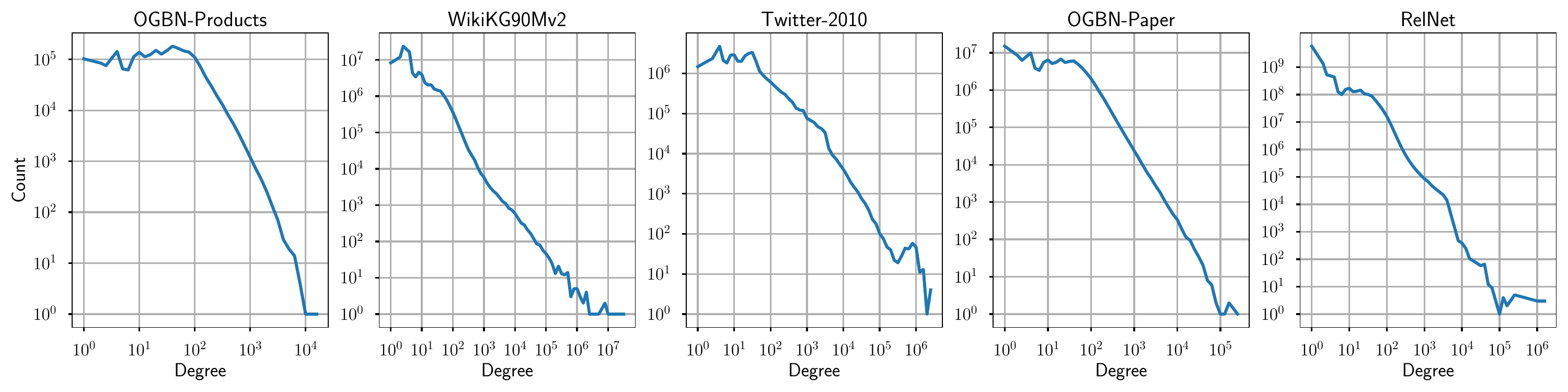}
  \caption{\label{fig:degree-hist} The degree distribution of datasets in log scale. }
\end{figure*}

\autoref{fig:degree-hist} shows the degree distribution of these datasets in log scale. It can be seen that all datasets except OGBN-Products follow a power law distribution.

\subsection{Graph Partitioner}
\label{sec:eval-graph-partition}

We run ParMETIS, DistributedNE, and AdaDNE on each of the 5 datasets mentioned above and evaluate their quality metrics, i.e., $RF$, $VB$, $EB$, and the runtime. Two different partition numbers are chosen according to the dataset size and the partition results are listed in \autoref{table:gp}. The default parameter settings are used for ParMETIS and \dne. For AdaDNE, we set $\lambda_p^0$ to \dne's default expansion factor of 0.1, and $\alpha=\beta=1$.

\renewcommand{\arraystretch}{1}
\begin{table*}[!htbp]
  \centering
  \caption{Performance of different graph partition algorithms.}
  \label{table:gp}
  \begin{tabular}{>{\centering}p{0.25\columnwidth}|>{\centering}p{0.2\columnwidth}|>{\centering}p{0.25\columnwidth}|>{\centering}p{0.15\columnwidth}|>{\centering}p{0.15\columnwidth}|>{\centering}p{0.15\columnwidth}|>{\centering}p{0.15\columnwidth}}
    \hline
    Dataset&\# Partitions&Algorithm&$RF$&$VB$&$EB$&Time(s)\tabularnewline \hline
    \multirow{6}*{\ogbp}&\multirow{3}*{2}&ParMETIS&1.252&1.119&1.072&281 \tabularnewline \cline{3-7}
                                                &&DistributedNE&\bf{1.218}&1.129&1.036&\bf{148} \tabularnewline \cline{3-7}
                                                &&AdaDNE&1.389&\bf{1.060}&\bf{1.020}&234 \tabularnewline \cline{2-7}
                               &\multirow{3}*{4}&ParMETIS&\bf{1.494}&1.297&1.376&155 \tabularnewline \cline{3-7}
                                                &&DistributedNE&1.496&1.117&1.070&\bf{70} \tabularnewline \cline{3-7}
                                                &&AdaDNE&1.787&\bf{1.087}&\bf{1.053}&71 \tabularnewline \hline
    \multirow{6}*{\wiki}&\multirow{3}*{8}&ParMETIS&2.343&3.967&5.059&6515 \tabularnewline \cline{3-7}
                                              &&DistributedNE&\bf{1.276}&2.902&1.159&574 \tabularnewline \cline{3-7}
                                              &&AdaDNE&1.324&\bf{1.248}&\bf{1.083}&\bf{279} \tabularnewline \cline{2-7}
                             &\multirow{3}*{16}&ParMETIS&2.232&10.075&11.750&4598 \tabularnewline \cline{3-7}
                                              &&DistributedNE&\bf{1.327}&3.388&1.316&403 \tabularnewline \cline{3-7}
                                              &&AdaDNE&1.452&\bf{1.254}&\bf{1.140}&\bf{209} \tabularnewline \hline
    \multirow{6}*{\twitter}&\multirow{3}*{8}&ParMETIS&3.603&1.592&7.278&4737 \tabularnewline \cline{3-7}
                                           &&DistributedNE&\bf{1.552}&2.770&1.085&899 \tabularnewline \cline{3-7}
                                           &&AdaDNE&1.631&\bf{1.216}&\bf{1.035}&\bf{720} \tabularnewline \cline{2-7}
                           &\multirow{3}*{16}&ParMETIS&5.025&3.600&16.479&4246 \tabularnewline \cline{3-7}
                                            &&DistributedNE&\bf{1.900}&5.634&1.429&537 \tabularnewline \cline{3-7}
                                            &&AdaDNE&2.058&\bf{2.730}&\bf{1.186}&\bf{426} \tabularnewline \hline
    \multirow{6}*{\ogbpa}&\multirow{3}*{8}&ParMETIS&\bf{1.517}&1.214&3.576&5403 \tabularnewline \cline{3-7}
                                                &&DistributedNE&2.374&1.471&1.216&899 \tabularnewline \cline{3-7}
                                                &&AdaDNE&2.176&\bf{1.170}&\bf{1.021}&\bf{840} \tabularnewline \cline{2-7}
                               &\multirow{3}*{16}&ParMETIS&\bf{1.837}&1.731&4.094&4173 \tabularnewline \cline{3-7}
                                                &&DistributedNE&2.763&1.583&1.210&780 \tabularnewline \cline{3-7}
                                                &&AdaDNE&2.532&\bf{1.428}&\bf{1.045}&\bf{683} \tabularnewline \hline
    \multirow{6}*{RelNet}&\multirow{3}*{32}&ParMETIS&\multicolumn{4}{c}{OOM} \tabularnewline \cline{3-7}
                                           &&DistributedNE&\multicolumn{4}{c}{OOM} \tabularnewline \cline{3-7}
                                           &&AdaDNE&\bf{1.840}&\bf{1.121}&\bf{1.024}&\bf{33156} \tabularnewline \cline{2-7}
                           &\multirow{3}*{64}&ParMETIS&\multicolumn{4}{c}{OOM} \tabularnewline \cline{3-7}
                                           &&DistributedNE&\multicolumn{4}{c}{OOM} \tabularnewline \cline{3-7}
                                           &&AdaDNE&\bf{2.117}&\bf{1.252}&\bf{1.014}&\bf{22400} \tabularnewline \hline
\end{tabular}
\end{table*}

As shown in \autoref{table:gp}, The performance of edge-cut-based ParMETIS is notably inferior to vertex-cut based \dne and AdaDNE on power law graphs in terms of partitioning quality and elapsed time. This highlights that vertex-cut partitioning is more effective for power law graphs. AdaDNE's optimizations effectively reduce data skew, resulting in the lowest $VB$ and $EB$ metrics in all cases, as well as comparable $RF$ and elapsed time with DNE. Moreover, only AdaDNE can partition the ten billion scale RelNet dataset, as both ParMETIS and \dne suffer from OOM failures.

\subsection{Neighbor Sampling Performance}
\label{sec:eval-sampling}

We compared the uniform and weighted neighbor sampling performance of \sysname with DistDGL, PaGraph and GraphLearn. The RelNet dataset was excluded because none of these frameworks for comparison could handle such a large scale dataset. At the graph partitioning stage, \sysname and DistDGL/PaGraph adopt AdaDNE and ParMETIS, respectively. GraphLearn adopts Hash partitioning, which is the only partition algorithm it provides. The partition number is set to 2 for \ogbp and 8 for \wiki, \twitter and \ogbpa. The sampling fanouts are $[15, 10, 5]$, all vertices had their weights set to 1. We uniformly sample an equal number of vertices from each partition to build the seed set. This setup emulates the balanced seed design of DistDGL, where the seed vertices are expected to be evenly distributed across each partition. In addition, DistDGL requires the same number of clients and graph servers, we apply the same settings for \sysname and GraphLearn, i.e. 2 clients for \ogbp and 8 clients for \wiki and \twitter. The reported sampling speed is the \textbf{average} of all clients.

\begin{figure}[!htbp]
  \centering
    \includegraphics[width=\columnwidth]{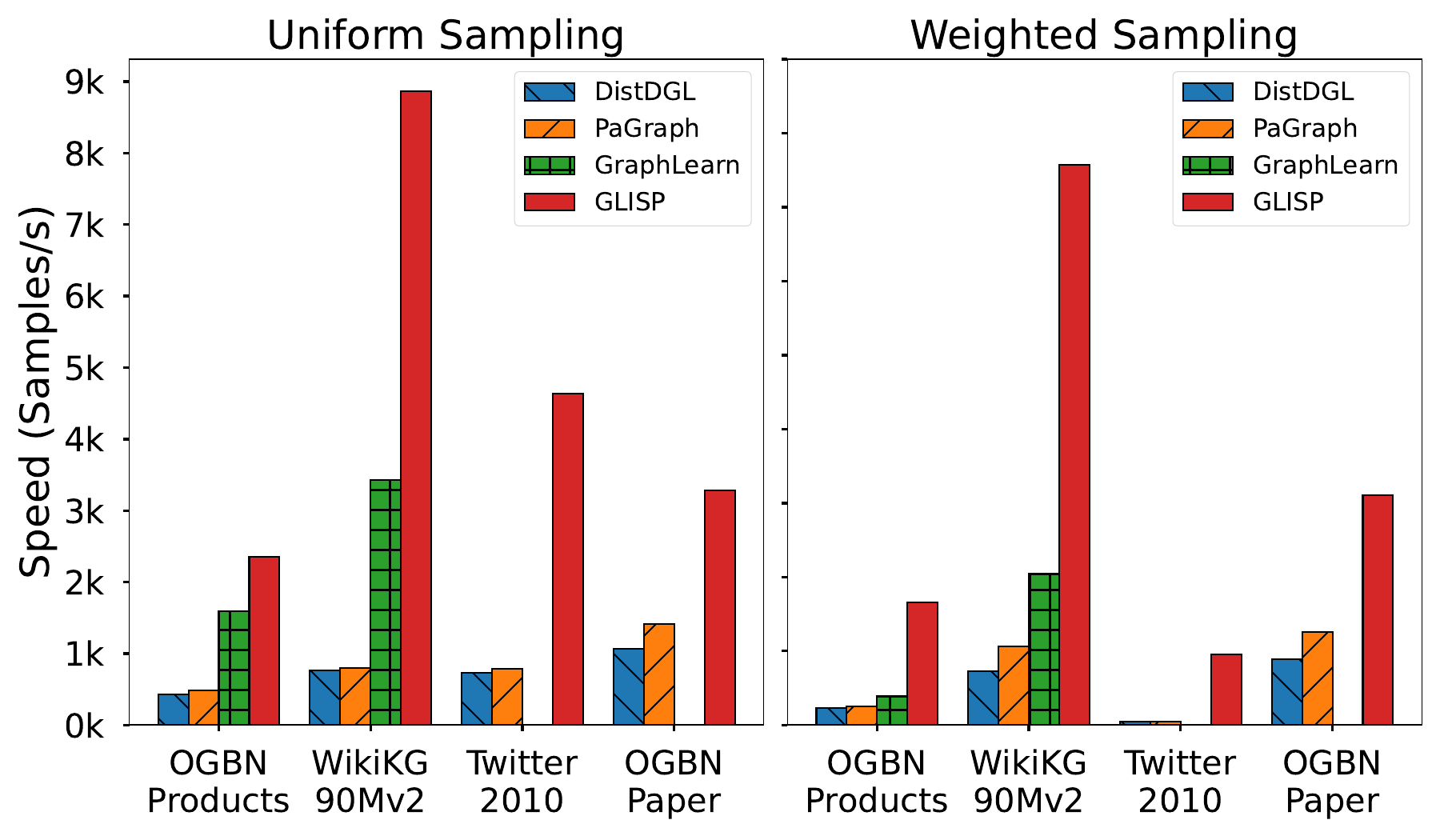}
  \caption{Subgraph sampling speed of different frameworks. GraphLearn fails on \twitter and \ogbpa due to OOM, so the corresponding result is not presented.\label{fig:sampling-speed}}
\end{figure}

\autoref{fig:sampling-speed} show that \sysname achieves the highest speed in all cases. We attribute the performance gain to two aspects: 1) the Gather-Apply paradigm based sampling architecture in conjunction with the high-quality AdaDNE algorithm results in a balanced workload, 2) the contiguous memory layout provides higher access speed. In addition, the speedup of \sysname is more pronounced in weighted sampling settings, because the impact of workload imbalances is amplified by complicated weighted sampling algorithms. We note that GraphLearn fails on the \twitter dataset due to OOM, highlighting the poor scalability of the Hash partitioning and data structure of GraphLearn.

\begin{figure}[!htbp]
  \centering
  \includegraphics[width=\columnwidth]{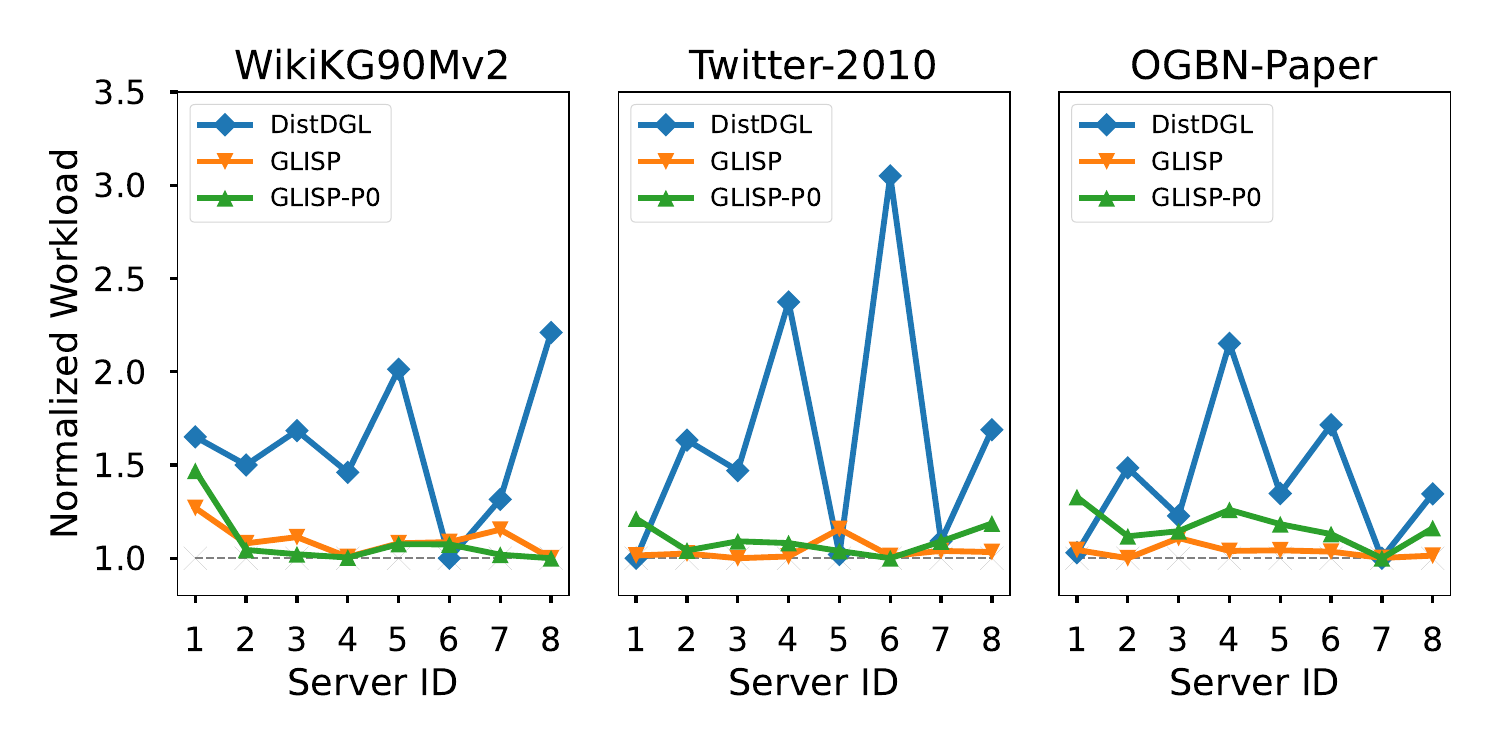}
  \caption{Server workload of DistDGL and \sysname. The curve shown in the legend \sysname-P0 represents the worst case where all seed vertices are from partition 0 of \sysname.  We exclude the \ogbp dataset since the partition number is too small to be representative. \label{fig:load-balance}}
\end{figure}

To confirm the effectiveness of our load balancing design, which relies on vertex-cut partitioning and the Gather-Apply paradigm, we calculated the server workload of DistDGL and \sysname and present the results in \autoref{fig:load-balance}. The workload is measured in terms of throughput and normalized by the minimum throughput in the server group to eliminate discrepancies in raw data induced by framework implementations. Specifically, the normalized load $\bar{W_i}$ of server $i$ is defined by $\bar{W_i} = W_i/\min_{p\in \{P\}}(W_p)$, where $W_i$ is the raw throughout of server $i$ and $P$ is the total number of servers.

\autoref{fig:load-balance} show that although the seed vertices are evenly distributed across the servers, DistDGL still suffers from severe load imbalance, while the load of \sysname is well balanced. This verifies that simply balancing the seed vertices does not eliminate the data skew originating from graph partitioning. Furthermore, we conduct a worst-case test where all seeds are present on the first partition for \sysname, and show the result in \autoref{fig:load-balance}, legend \sysname-P0. The results show that the slight increased workload on server 0 degrades the load balancing of \sysname, but still significantly outperforms DistDGL. Notably, balancing the seed setting is only relevant in supervised vertex classification scenarios where seed vertices are explicitly provided.  In the self-supervised learning and graph inference scenarios, all vertices are considered as seeds, so the seed balance is equivalent to the vertex balance.

The server memory footprint of the different frameworks is given in \autoref{table:memory-footprint}. To remove the data redundancy introduced by different graph partition algorithms, we load the original graph directly on a machine with 256GB of memory. \sysname has the smallest memory footprint of all datasets, indicating that our data structure is more efficient and can handle larger graphs with given resources.

\renewcommand{\arraystretch}{1}
\begin{table}[!htbp]
  \centering
  \caption{Memory footprint of different frameworks in GB. We omit PaGraph as it is built on top of DistDGL and shares the same graph data structure.}
  \label{table:memory-footprint}
  \begin{tabular}{>{\centering}p{0.25\columnwidth}|>{\centering}p{0.2\columnwidth}|>{\centering}p{0.2\columnwidth}|>{\centering}p{0.2\columnwidth}}
    \hline
    Dataset&DistDGL&GraphLearn&\sysname \tabularnewline \hline
    \ogbp&2.0&5.5&\bf{0.6} \tabularnewline  \hline
    \wiki&18.0&79.6&\bf{10.9} \tabularnewline  \hline
    \twitter&21.9&67.4&\bf{12.7} \tabularnewline  \hline
    \ogbpa&24.1&84.2&\bf{16.8} \tabularnewline  \hline
\end{tabular}
\end{table}

\subsection{Model convergence and scalability}
\label{sec:eval-convergence-scalability}
To verify the correctness of \sysname, we evaluate the inductive vertex classification task on GCN\cite{kipfSemiSupervisedClassificationGraph2016}, GraphSAGE\cite{hamiltonInductiveRepresentationLearning2018} and GAT\cite{velickovicGraphAttentionNetworks2018} with \ogbp and \ogbpa datasets and compare the test accuracy with DistDGL, PaGraph and GraphLearn. The model implementations provided publicly by DistDGL were utilized, with the number of stacked GNN layers fixed at 3. The hidden size of the three models is set to 256 and 4 attention heads are used in GAT. The input subgraph samples are generated with fanouts $[15, 10, 5]$.

\begin{figure}[!htbp]
  \centering
  \includegraphics[width=\columnwidth]{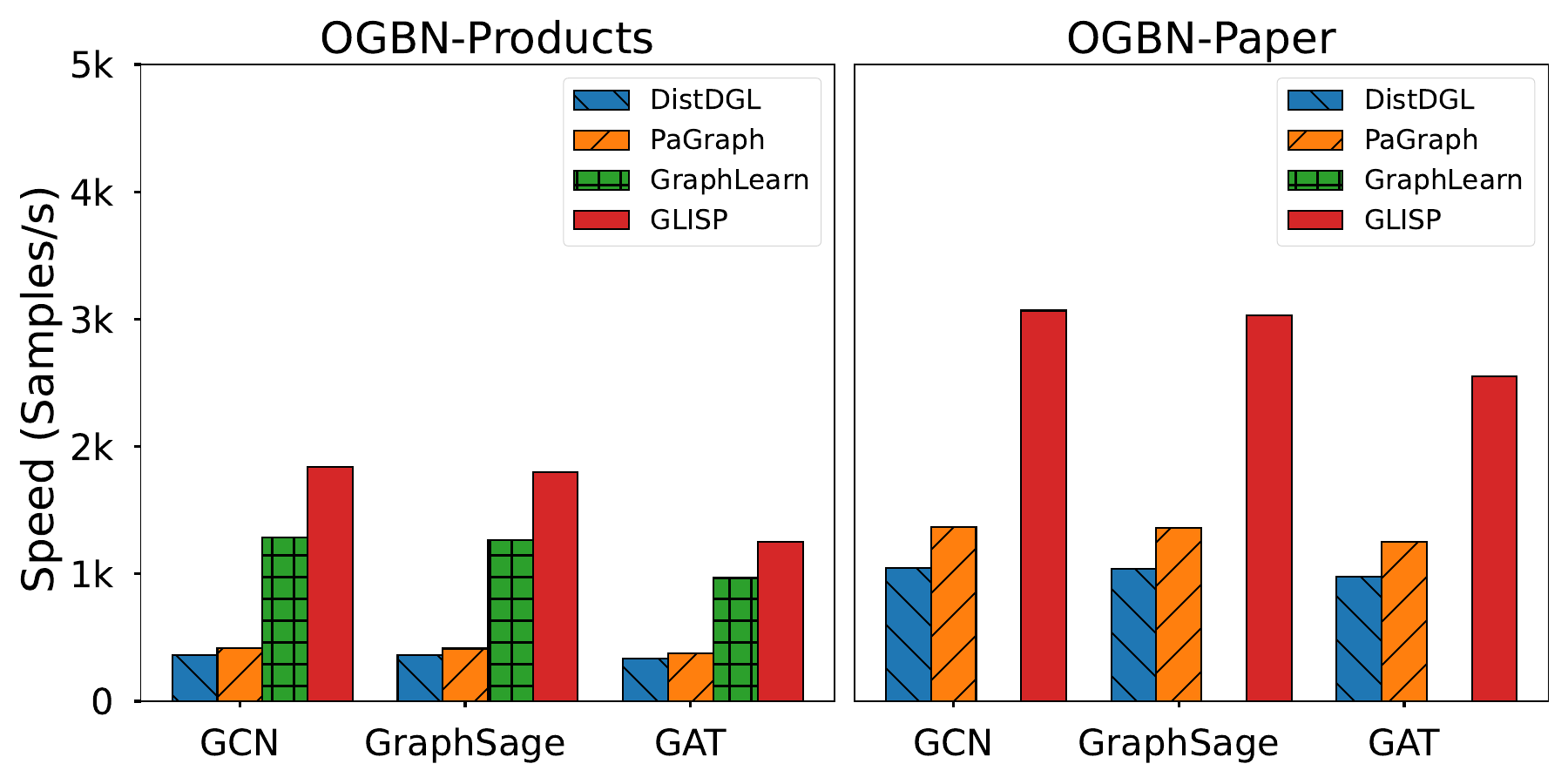}
  \caption{The end-to-end training speed of DistDGL, PaGraph, GraphLearn and \sysname. \label{fig:training-speed}}
\end{figure}

As shown in \autoref{table:test-acc}, the test accuracies of \sysname agree well with DistDGL and GraphLern. \autoref{fig:training-speed} shows the end-to-end training speed of different frameworks. \sysname achieves a speedup of $1.57\times\sim 6.53\times$ over DistDGL and GraphLern with the three models, which can be attributed to our load-balanced graph sampling service. 

\renewcommand{\arraystretch}{1}
\begin{table}[!htbp]
  \centering
  \caption{Test accuracy of different frameworks on 3 GNN models.}
  \label{table:test-acc}
  \begin{tabular}{>{\centering}p{0.15\columnwidth}|>{\centering}p{0.2\columnwidth}|>{\centering}p{0.15\columnwidth}|>{\centering}p{0.17\columnwidth}|>{\centering}p{0.15\columnwidth}}
    \hline
    DataSet&Framework&GCN&GraphSAGE&GAT \tabularnewline \hline
    \multirow{2}*{\makecell{OGBN-\\Products}}&DistDGL&0.763&0.775&0.759 \tabularnewline 
    &PaGraph&0.761&0.773&0.761 \tabularnewline          
    &GraphLearn&0.762&0.775&0.760 \tabularnewline       
    &\sysname&0.760&0.774&0.762 \tabularnewline       
    \hline
    \multirow{2}*{\makecell{OGBN-\\Paper}}&DistDGL&0.455&0.459&0.470 \tabularnewline 
    &PaGraph&0.456&0.460&0.474 \tabularnewline          
    &GraphLearn&0.452&0.459&0.471 \tabularnewline       
    &\sysname&0.451&0.457&0.475 \tabularnewline       
    \hline
\end{tabular}
\end{table}

\begin{figure}[!htbp]
  \centering
  \includegraphics[width=\columnwidth]{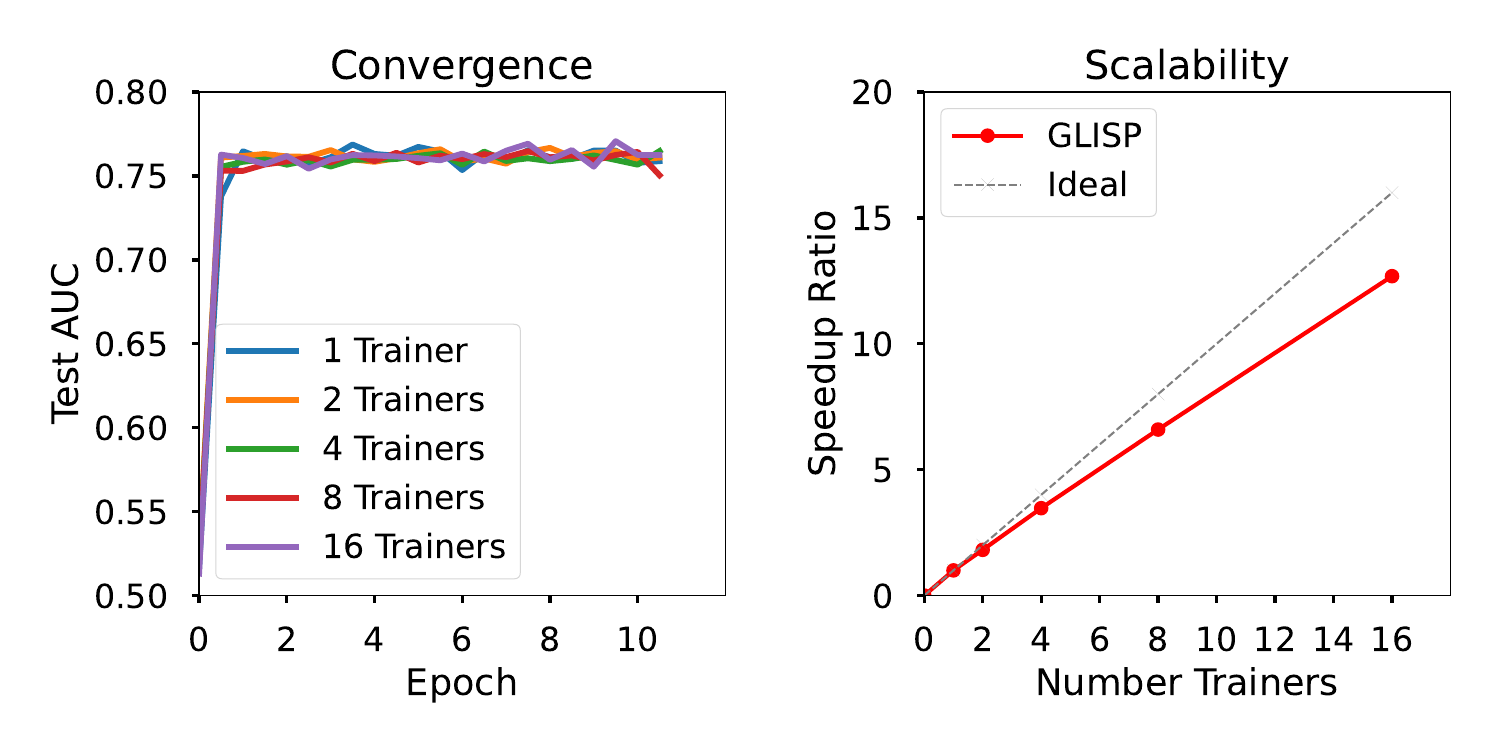}
  \caption{Convergence and scalability of \sysname.\label{fig:scalability}}
\end{figure}

We train a Knowledge Graph Embedding(KGE) model\cite{wangKnowledgeGraphEmbedding2017} on the large scale dataset RelNet(32 partitions) to verify the scalability of \sysname. The encoder is 2 layer Heterogeneous Graph Transformer(HGT)\cite{huHeterogeneousGraphTransformer2020} with the hidden size of 128 and the decoder is 2 layer feed forward neural network. 100M edges are randomly selected as positive samples of training set and negative samples are generated by replacing the head or tail with random vertex. We follow the synchronous training approach, where increasing the number of trainers is equivalent to increasing the batch size. \autoref{fig:scalability} (a) is the convergence curve, revealing that the number of trainers does not influence the model performance. \autoref{fig:scalability} (b) shows the speedup ratio, where the gray dotted line represents the ideal speedup ratio with slope of 1, and the red solid line is the speedup achieved by \sysname, with a slope of about 0.8.

\subsection{Graph Inference Engine}
\label{sec:eval-full-graph-inference}

We validate the performance of the graph inference engine with the aforementioned KGE model trained on the 32 partition RelNet dataset. The experiment contains two tasks: \textbf{vertex embedding}  and \textbf{link prediction}, where the former produces a 128 dimensional embedding for each vertex, and the latter predicts the score for each edge. Both tasks run on 32 GPU workers, each for one partition, and the intermediate embeddings are stored on our internal HDFS cluster.

\begin{figure}[!htbp]
  \centering
  \includegraphics[width=0.7\columnwidth]{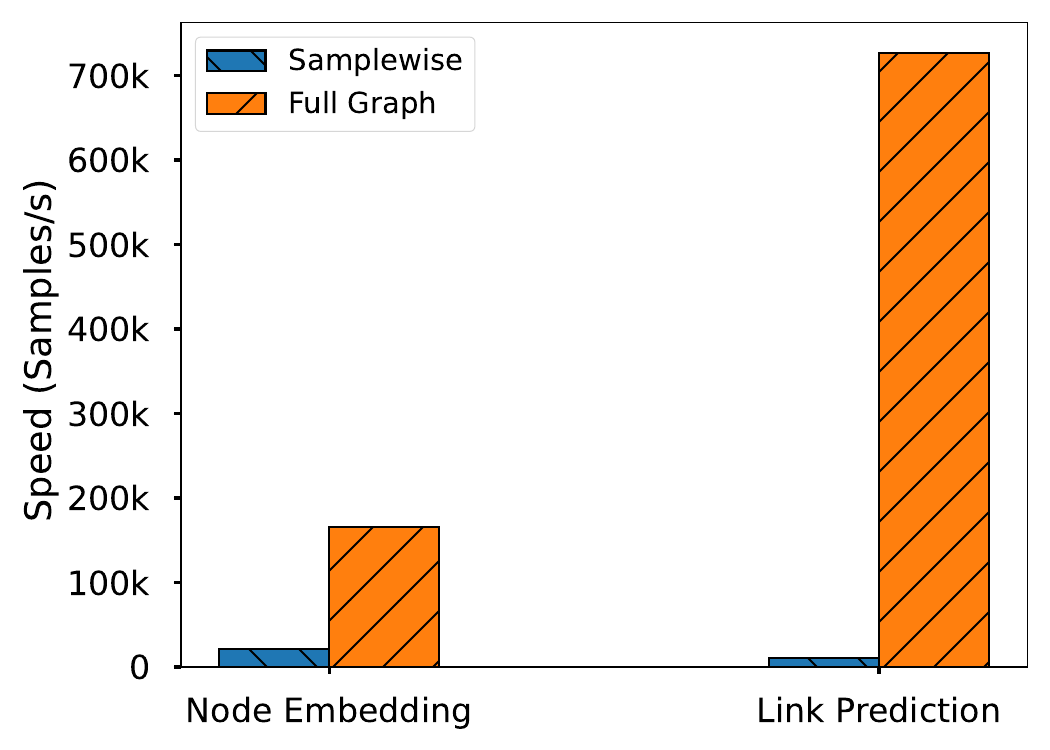}
  \caption{Overall performance of full graph inference versus naive samplewise inference. The term "naive" indicates that the inference simply follows the training mode without optimizations such as GNN slice, embedding cache and graph reorder. The full graph inference gains $7.89\times$ and $70.77\times$ speedup on the vertex embedding and link prediction tasks, respectively. \label{fig:inference-speed}}
\end{figure}

As shown in \autoref{fig:inference-speed}, the full graph inference obtains a $7.89\times$ and $70.77\times$ speedup over naive samplewise inference on the vertex embedding and link prediction tasks, requiring only 17.5h and 18.7h to complete the inference, respectively. In addition, we note that the link prediction task exhibits a higher speedup ratio compared to the vertex embedding task, which is due to the fact that the edge prediction score depends on the embedding of both endpoints, thus exacerbating the redundant computation issue of samplewise inference.

As a key design of the graph inference engine, the caching system has a significant impact on the overall performance. We conduct a series of  drill-down experiments to verify the effectiveness of the two-level cache and the PDS algorithm in the caching system.  Three graph reorder algorithms are chosen for comparison with our PDS, namely Natural Sort (NS),  Degree Sort (DS), and Partition Sort (PS). Essentially, these four algorithms are equivalent to sorting vertices by different keys. In particular, the keys of NS, DS, PS and PDS algorithm are $global\_id$, $degree$, $(partition\_id, global\_id)$ and $(partition\_id, degree)$, respectively.

\begin{figure}[!htbp]
  \centering
    \includegraphics[width=\columnwidth]{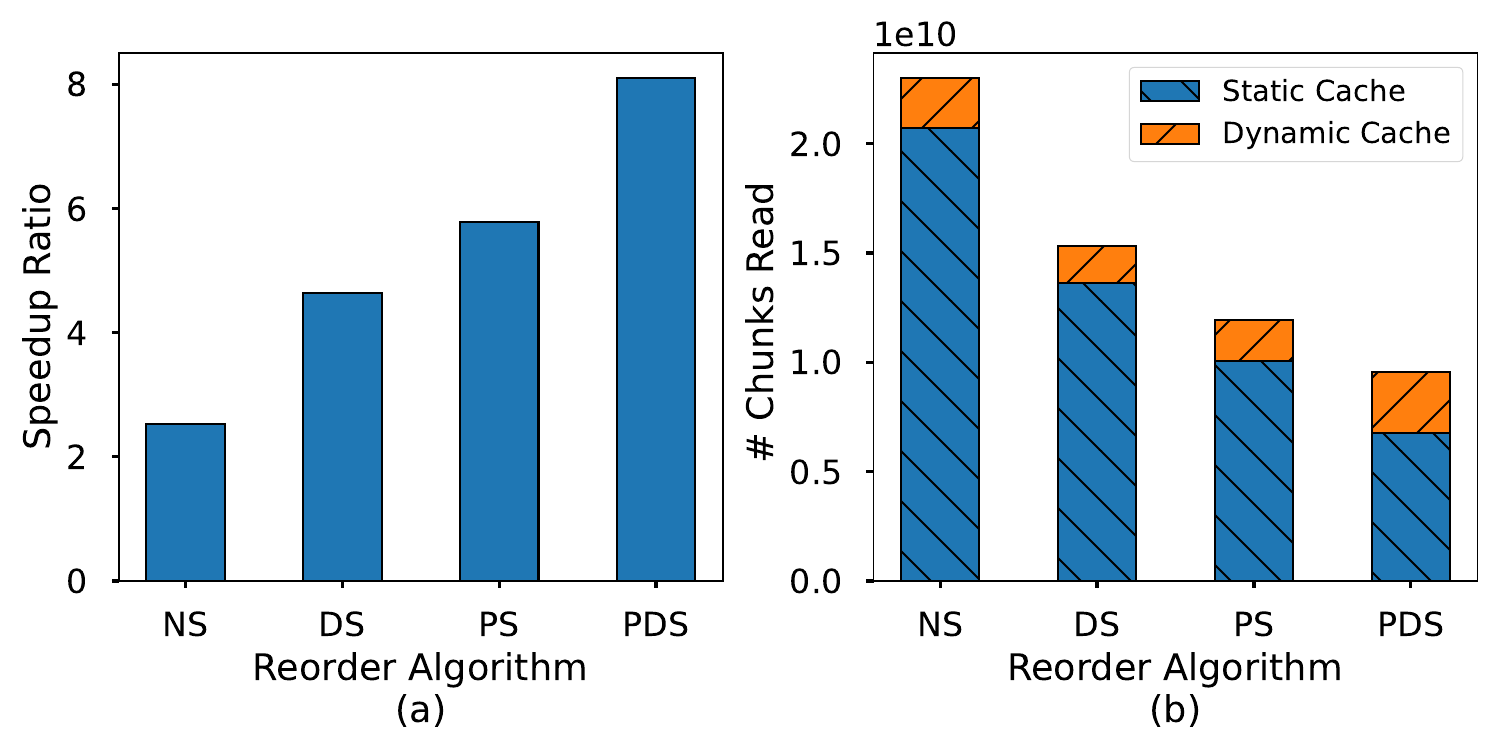}
  \caption{(a). speedup ratio and (b). total number of chunks read from the caching system of different graph reorder algorithms.\label{fig:cache-speedup-and-chunks-read}}
\end{figure}

\autoref{fig:cache-speedup-and-chunks-read}(a) is the speedup ratio of fetching embedding through the caching system under different graph reorder algorithms with the baseline of reading directly from HDFS. We observe a $2.52\times$ speedup(NS) without graph reordering, and the speedup ratio can be further improved by graph reordering. This is due to the impact of the graph reorder algorithm on both the total number of chunks read and the hit rate of the high-speed memory cache.

\autoref{fig:cache-speedup-and-chunks-read}(b) further gives the total number of chunks to be read from the caching system in a full graph inference task, which consists of the number of reads from the static cache and the dynamic cache. We set the embedding chunk size to 32768. The dynamic cache size is $10\%$ of the number of chunks in each worker, and is updated with FIFO policy. By performing secondary reordering on the partitioned graph, the PDS algorithm fully exploits the locality of the graph data, achieves the lowest chunk reads (only 41.5\% of NS) and the highest dynamic cache hit ratio (over 29\%) simultaneously, thus outperforming other algorithms with a speedup ratio of 8.103. Furthermore, DS performs worse than PS, because the lightweight DS algorithm discards the locality that has already been identified by the graph partitioner.

\begin{figure}[!htbp]
  \centering
  \includegraphics[width=\columnwidth]{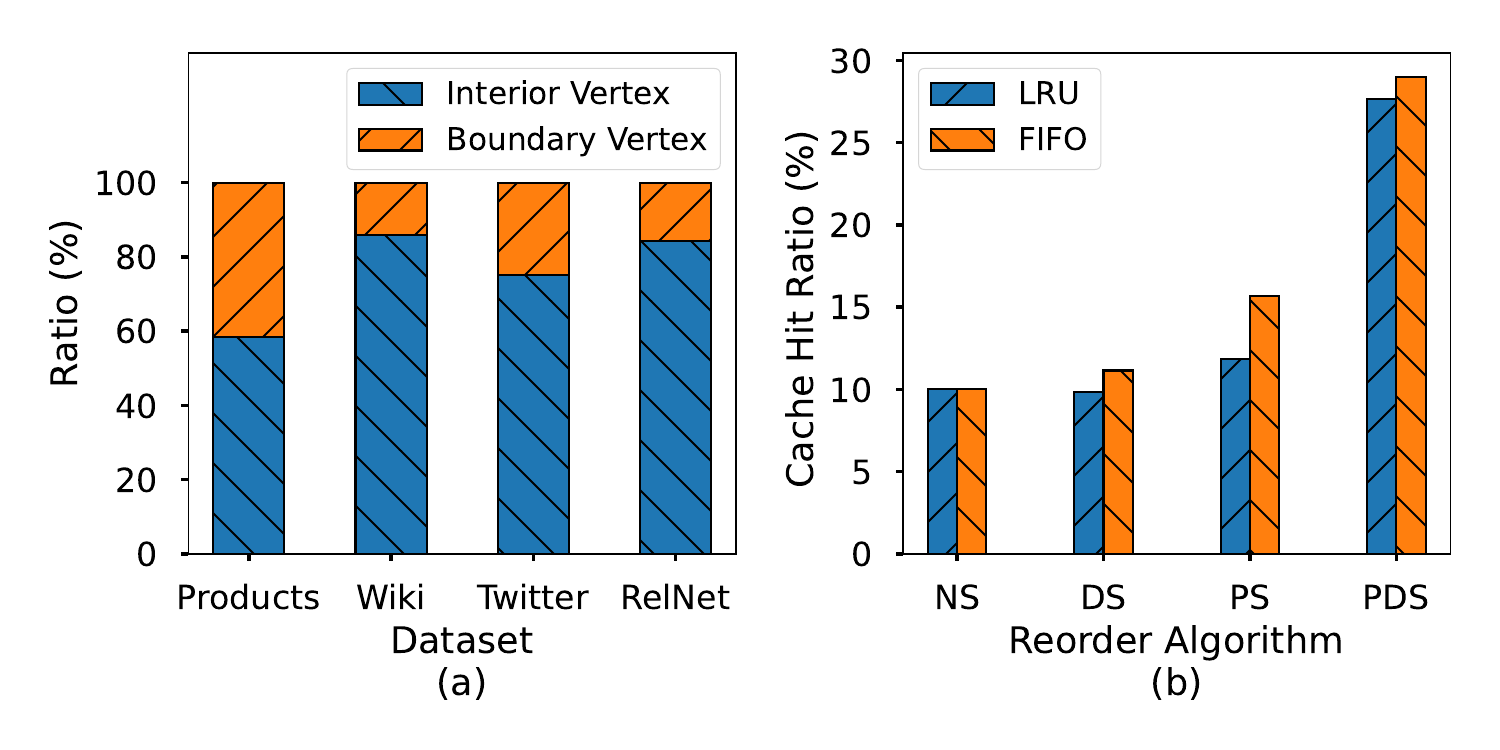}
  \caption{(a). The percentage of interior and boundary vertices in AdaDNE partitioned graph for \ogbp(Products), \wiki(Wiki), \twitter(Twitter) and RelNet, (b). Cache hit ratio for LRU and FIFO policies. Note that the absolute value of the hit ratio is impaired by the sparsity of industrial dataset RelNet, since the overlap between subgraphs is not as high as in the academic dense graphs.  \label{fig:interior-node-ratio-and-cache-hit-ratio}}
\end{figure}

The time cost of static cache filling and model inference is listed in \autoref{table:time-cost}, revealing that cache filling requires significantly less time than model inference($<10\%$). \autoref{fig:interior-node-ratio-and-cache-hit-ratio}(a) shows the percentage of interior and boundary vertices in the AdaDNE partitioned graph for different datasets, where the number of partitions is 2, 8, 8, and 32 for \ogbp, \wiki, \twitter and RelNet, respectively. It can be seen that the majority of the vertices are interior vertices, especially in the power law graphs where interior vertices constitute over $75\%$, which proves the effectiveness of AdaDNE algorithm and justifies our partition based static cache design.

\renewcommand{\arraystretch}{1}
\begin{table}[!htbp]
  \centering
  \caption{Time cost of filling the static cache and model inference.}
  \label{table:time-cost}
  \begin{tabular}{>{\centering}p{0.25\columnwidth}|>{\centering}p{0.3\columnwidth}|>{\centering}p{0.3\columnwidth}}
    \hline
    Task&Fill Cache Time (s)&Model Time (s) \tabularnewline \hline
    Vertex Embedding&3251&59987 \tabularnewline \hline
    Link Prediction&5635&61760 \tabularnewline 
    \hline
\end{tabular}
\end{table}

Regarding the update policy of the dynamic cache, we test two polices of LRU and FIFO, and show the hit ratio in \autoref{fig:interior-node-ratio-and-cache-hit-ratio}(b).  As LRU proved to be more intricate than FIFO and does not result in a higher hit ratio, we opted for the FIFO policy for the dynamic cache.

\section{Related Work}
\subsection{Graph Partitioning}
The most widely used edge-cut partitioning algorithm is  METIS\cite{karypisFastHighQuality} and its parallel version ParMETIS\cite{karypisParallelAlgorithmMultilevel1998}. METIS first coarsens the original graph $G$ into a series of smaller sized subgraphs $\{G_0,\dots G_n\}$ by collapsing vertices and edges, and then performs K-way partition on $G_n$ to divide it into $K$ parts. Finally the partitioned subgraphs are uncoarsened to construct partitioning of the original graph $G$.

\dne\cite{hanaiDistributedEdgePartitioning2019} is a SOTA vertex-cut partitioning algorithm for trillion-edge graphs. \dne adopts a neighbor expansion process to partition the graph, where each partition is constructed in parallel by greedily expanding its edge set from randomly selected seed vertices. In order to obtain balanced partitioning, \dne sets a threshold $E_{t}=\tau\frac{|E|}{|P|}$ for $|E_p|$, where $\tau$ is the imbalance factor. The expansion process of the $p$th partition will terminate when the number of edges $|E_p|$ satisfies $|E_p|>E_t$.

Recently, several GNN frameworks have implemented edge-cut based algorithms optimized for GNN training. DistDGL\cite{zhengDistDGLDistributedGraph2020} introduces constraints to balance the training set in ParMETIS. BGL\cite{liuBGLGPUEfficientGNN2021} and ByteGNN\cite{zhengByteGNNEfficientGraph} employ a block-based strategy and also incorporate constraints to balance the training set.

\subsection{Distributed GNN Systems}
Several distributed systems have been developed to facilitate GNN learning tasks on large graph data. ROC\cite{ImprovingAccuracyScalability}, PipeGCN\cite{wanPipeGCNEfficientFullGraph2021}, CAGNET\cite{tripathyReducingCommunicationGraph2020} are full-batch training systems that require no neighbor sampling. AGL\cite{zhangAGLScalableSystem2020} and GDLL\cite{thuvanGDLLScalableShare2022} employ an \textbf{offline} sampling strategy where the $K$ hop subgraphs of each vertex are sampled and stored on disk via MapReduce prior to the training. However, the sampling process is time-intensive and the size of $K$ hop subgraphs grows exponentially as $K$ increases, which is a major challenge for both storage and network traffic. To address these challenges, DistDGL\cite{ zhengDistDGLDistributedGraph2020, zhengDistributedHybridCPU2022} and GraphLearn\cite{zhuAliGraphComprehensiveGraph2019} propose an \textbf{online} graph sampling service that performs sampling during training. A number of systems are built on top of DGL or GraphLearn and introduce several training optimizations, such as PaGraph\cite{linPaGraphScalingGNN2020}, BGL\cite{liuBGLGPUEfficientGNN2021} and ByteGNN\cite{zhengByteGNNEfficientGraph}. DGI\cite{yinDGIEasyEfficient2023}, also based on DGL, proposes a layerwise inference strategy that can automatically partitions computational graphs into layers. However, DGI operates on a single machine and lacks a targeted design for distributed environments, including graph partitioning and distributed embedding retrieval, making it difficult to scale to industrial graphs.

\section{Conclusion}
\label{sec:conclusion}
We present \sysname, an structural property aware online GNN learning framework designed for industrial scale graphs. \sysname is composed of graph partitioner, graph sampling service and graph inference engine. We propose a vertex-cut based graph partition algorithm AdaDNE for the graph partitioner, which can better handle power law graphs and reduce data skew and redundancy. Based on our compact and memory efficient data structure for vertex-cut partitioned graphs, the graph sampling service implements load balanced uniform/weighted neighbor sampling algorithms in a Gather-Apply fasion. The graph inference engine follows the message passing paradigm and perform inference layer by layer. The intermediate embeddings of each layer are stored on a distributed file system (DFS) to improve scalability. To reduce the embedding retrieval time, the graph inference engine incorporates a two-level embedding caching system. By leveraging the data locality mined by our proposed graph reordering algorithm PDS, the caching system significantly improves the embedding retrieval speed. Experiments show that \sysname achieves a $1.57\times\sim 6.53\times$ speedup in training and up to $70.77\times$ speedup in inference compared to the SOTA frameworks.

\newpage

\bibliography{main}

\begin{thebibliography}{10}
\providecommand{\url}[1]{#1}
\csname url@samestyle\endcsname
\providecommand{\newblock}{\relax}
\providecommand{\bibinfo}[2]{#2}
\providecommand{\BIBentrySTDinterwordspacing}{\spaceskip=0pt\relax}
\providecommand{\BIBentryALTinterwordstretchfactor}{4}
\providecommand{\BIBentryALTinterwordspacing}{\spaceskip=\fontdimen2\font plus
\BIBentryALTinterwordstretchfactor\fontdimen3\font minus
  \fontdimen4\font\relax}
\providecommand{\BIBforeignlanguage}[2]{{%
\expandafter\ifx\csname l@#1\endcsname\relax
\typeout{** WARNING: IEEEtran.bst: No hyphenation pattern has been}%
\typeout{** loaded for the language `#1'. Using the pattern for}%
\typeout{** the default language instead.}%
\else
\language=\csname l@#1\endcsname
\fi
#2}}
\providecommand{\BIBdecl}{\relax}
\BIBdecl

\bibitem{wangKnowledgeGraphEmbedding2017}
Q.~Wang, Z.~Mao, B.~Wang, and L.~Guo, ``Knowledge graph embedding: A survey of
  approaches and applications,'' \emph{IEEE Transactions on Knowledge and Data
  Engineering}, vol.~29, no.~12, pp. 2724--2743, 2017.

\bibitem{donatoLargeScaleProperties2004a}
D.~Donato, L.~Laura, S.~Leonardi, and S.~Millozzi, ``Large scale properties of
  the webgraph,'' \emph{The European Physical Journal B}, vol.~38, pp.
  239--243, 2004.

\bibitem{yingGraphConvolutionalNeural2018}
R.~Ying, R.~He, K.~Chen, P.~Eksombatchai, W.~L. Hamilton, and J.~Leskovec,
  ``Graph convolutional neural networks for web-scale recommender systems,'' in
  \emph{Proceedings of the 24th ACM SIGKDD international conference on
  knowledge discovery \& data mining}, 2018, pp. 974--983.

\bibitem{kipfSemiSupervisedClassificationGraph2016}
T.~N. Kipf and M.~Welling, ``Semi-supervised classification with graph
  convolutional networks,'' \emph{arXiv preprint arXiv:1609.02907}, 2016.

\bibitem{hamiltonInductiveRepresentationLearning2018}
W.~Hamilton, Z.~Ying, and J.~Leskovec, ``Inductive representation learning on
  large graphs,'' \emph{Advances in neural information processing systems},
  vol.~30, 2017.

\bibitem{liuGeniePathGraphNeural2018}
Z.~Liu, C.~Chen, L.~Li, J.~Zhou, X.~Li, L.~Song, and Y.~Qi, ``Geniepath: Graph
  neural networks with adaptive receptive paths,'' in \emph{Proceedings of the
  AAAI Conference on Artificial Intelligence}, vol.~33, no.~01, 2019, pp.
  4424--4431.

\bibitem{shumanEmergingFieldSignal2013}
D.~I. Shuman, S.~K. Narang, P.~Frossard, A.~Ortega, and P.~Vandergheynst, ``The
  emerging field of signal processing on graphs: Extending high-dimensional
  data analysis to networks and other irregular domains,'' \emph{IEEE signal
  processing magazine}, vol.~30, no.~3, pp. 83--98, 2013.

\bibitem{velickovicGraphAttentionNetworks2018}
P.~Veli{\v{c}}kovi{\'c}, G.~Cucurull, A.~Casanova, A.~Romero, P.~Li{\`o}, and
  Y.~Bengio, ``Graph attention networks,'' in \emph{International Conference on
  Learning Representations}, 2018.

\bibitem{zhouGraphNeuralNetworks2019}
J.~Zhou, G.~Cui, S.~Hu, Z.~Zhang, C.~Yang, Z.~Liu, L.~Wang, C.~Li, and M.~Sun,
  ``Graph neural networks: A review of methods and applications,'' \emph{AI
  Open}, vol.~1, pp. 57--81, 2020.

\bibitem{Jizhe2018Billion}
J.~W. H. Z. Z. Z.~L. Lee, ``Billion-scale commodity embedding for e-commerce
  recommendation in alibaba,'' \emph{SIGKDD explorations}, no. Udisk, 2018.

\bibitem{zengGraphSAINTGraphSampling2020}
H.~Zeng, H.~Zhou, A.~Srivastava, R.~Kannan, and V.~Prasanna, ``Graphsaint:
  Graph sampling based inductive learning method,'' in \emph{International
  Conference on Learning Representations}, 2019.

\bibitem{chiangClusterGCNEfficientAlgorithm2019}
W.~L. Chiang, X.~Liu, S.~Si, Y.~Li, and C.~J. Hsieh, ``Cluster-gcn: An
  efficient algorithm for training deep and large graph convolutional
  networks,'' \emph{ACM}, 2019.

\bibitem{qiuGCCGraphContrastive2020}
J.~Qiu, Q.~Chen, Y.~Dong, J.~Zhang, H.~Yang, M.~Ding, K.~Wang, and J.~Tang,
  ``Gcc: Graph contrastive coding for graph neural network pre-training,'' in
  \emph{Proceedings of the 26th ACM SIGKDD International Conference on
  Knowledge Discovery \& Data Mining}, 2020, pp. 1150--1160.

\bibitem{wangDeepGraphLibrary2020}
M.~Wang, D.~Zheng, Z.~Ye, Q.~Gan, M.~Li, X.~Song, J.~Zhou, C.~Ma, L.~Yu, Y.~Gai
  \emph{et~al.}, ``Deep graph library: A graph-centric, highly-performant
  package for graph neural networks,'' \emph{arXiv preprint arXiv:1909.01315},
  2019.

\bibitem{feyFastGraphRepresentation2019}
M.~Fey and J.~E. Lenssen, ``Fast graph representation learning with pytorch
  geometric,'' \emph{arXiv preprint arXiv:1903.02428}, 2019.

\bibitem{zhengDistDGLDistributedGraph2020}
D.~Zheng, C.~Ma, M.~Wang, J.~Zhou, Q.~Su, X.~Song, Q.~Gan, Z.~Zhang, and
  G.~Karypis, ``Distdgl: distributed graph neural network training for
  billion-scale graphs,'' in \emph{2020 IEEE/ACM 10th Workshop on Irregular
  Applications: Architectures and Algorithms (IA3)}.\hskip 1em plus 0.5em minus
  0.4em\relax IEEE, 2020, pp. 36--44.

\bibitem{zhengDistributedHybridCPU2022}
D.~Zheng, X.~Song, C.~Yang, D.~LaSalle, and G.~Karypis, ``Distributed hybrid
  cpu and gpu training for graph neural networks on billion-scale heterogeneous
  graphs,'' in \emph{Proceedings of the 28th ACM SIGKDD Conference on Knowledge
  Discovery and Data Mining}, 2022, pp. 4582--4591.

\bibitem{zhuAliGraphComprehensiveGraph2019}
R.~Zhu, K.~Zhao, H.~Yang, W.~Lin, C.~Zhou, B.~Ai, Y.~Li, and J.~Zhou,
  ``Aligraph: A comprehensive graph neural network platform,''
  \emph{Proceedings of the VLDB Endowment}, vol.~12, no.~12.

\bibitem{EulerGithub2022}
\BIBentryALTinterwordspacing
``Euler github,'' {Alibaba}. [Online]. Available:
  \url{https://github.com/alibaba/euler}
\BIBentrySTDinterwordspacing

\bibitem{liuBGLGPUEfficientGNN2021}
\BIBentryALTinterwordspacing
T.~Liu, Y.~Chen, D.~Li, C.~Wu, Y.~Zhu, J.~He, Y.~Peng, H.~Chen, H.~Chen, and
  C.~Guo, ``{BGL}: {GPU-Efficient} {GNN} training by optimizing graph data
  {I/O} and preprocessing,'' in \emph{20th USENIX Symposium on Networked
  Systems Design and Implementation (NSDI 23)}.\hskip 1em plus 0.5em minus
  0.4em\relax Boston, MA: USENIX Association, Apr. 2023, pp. 103--118.
  [Online]. Available:
  \url{https://www.usenix.org/conference/nsdi23/presentation/liu-tianfeng}
\BIBentrySTDinterwordspacing

\bibitem{zhengByteGNNEfficientGraph}
C.~Zheng, H.~Chen, Y.~Cheng, Z.~Song, Y.~Wu, C.~Li, J.~Cheng, H.~Yang, and
  S.~Zhang, ``Bytegnn: efficient graph neural network training at large
  scale,'' \emph{Proceedings of the VLDB Endowment}, vol.~15, no.~6, pp.
  1228--1242, 2022.

\bibitem{liDeeperGCNAllYou2020}
G.~Li, C.~Xiong, A.~Thabet, and B.~Ghanem, ``Deepergcn: All you need to train
  deeper gcns,'' \emph{arXiv preprint arXiv:2006.07739}, 2020.

\bibitem{liDeeperInsightsGraph2018}
Q.~Li, Z.~Han, and X.-M. Wu, ``Deeper insights into graph convolutional
  networks for semi-supervised learning,'' in \emph{Proceedings of the AAAI
  conference on artificial intelligence}, vol.~32, no.~1, 2018.

\bibitem{chenMeasuringRelievingOversmoothing2019}
D.~Chen, Y.~Lin, W.~Li, P.~Li, J.~Zhou, and X.~Sun, ``Measuring and relieving
  the over-smoothing problem for graph neural networks from the topological
  view,'' in \emph{Proceedings of the AAAI conference on artificial
  intelligence}, vol.~34, no.~04, 2020, pp. 3438--3445.

\bibitem{oonoGraphNeuralNetworks2021}
K.~Oono and T.~Suzuki, ``Graph neural networks exponentially lose expressive
  power for node classification,'' 2019.

\bibitem{caiNoteOverSmoothingGraph2020}
C.~Cai and Y.~Wang, ``A note on over-smoothing for graph neural networks,''
  2020.

\bibitem{karypisFastHighQuality}
G.~Karypis and V.~Kumar, ``A fast and high quality multilevel scheme for
  partitioning irregular graphs,'' \emph{SIAM Journal on scientific Computing},
  vol.~20, no.~1, pp. 359--392, 1998.

\bibitem{karypisParallelAlgorithmMultilevel1998}
------, ``A parallel algorithm for multilevel graph partitioning and sparse
  matrix ordering,'' \emph{Journal of parallel and distributed computing},
  vol.~48, no.~1, pp. 71--95, 1998.

\bibitem{slotaPartitioningTrillionedgeGraphs2016}
G.~M. Slota, S.~Rajamanickam, K.~Devine, and K.~Madduri, ``Partitioning
  trillion-edge graphs in minutes,'' in \emph{2017 IEEE International Parallel
  and Distributed Processing Symposium (IPDPS)}.\hskip 1em plus 0.5em minus
  0.4em\relax IEEE, 2017, pp. 646--655.

\bibitem{hanaiDistributedEdgePartitioning2019}
M.~Hanai, T.~Suzumura, W.~J. Tan, E.~Liu, G.~Theodoropoulos, and W.~Cai,
  ``Distributed edge partitioning for trillion-edge graphs,'' \emph{Proceedings
  of the VLDB Endowment}, vol.~12, no.~13.

\bibitem{kongClusteringbasedPartitioningLarge2022}
D.~Kong, X.~Xie, and Z.~Zhang, ``Clustering-based partitioning for large web
  graphs,'' \emph{arXiv preprint arXiv:2201.00472}, 2022.

\bibitem{cai2023dsp}
Z.~Cai, Q.~Zhou, X.~Yan, D.~Zheng, X.~Song, C.~Zheng, J.~Cheng, and G.~Karypis,
  ``Dsp: Efficient gnn training with multiple gpus,'' in \emph{Proceedings of
  the 28th ACM SIGPLAN Annual Symposium on Principles and Practice of Parallel
  Programming}, 2023, pp. 392--404.

\bibitem{crucittiErrorAttackTolerance2004}
R.~Albert, H.~Jeong, and A.~L. Barabasi, ``Error and attack tolerance of
  complex networks,'' \emph{Nature}, vol. 406, no. 6794, pp. 378--382, 2000.

\bibitem{donatoLargeScaleProperties2004}
D.~Donato, L.~Laura, S.~Leonardi, and S.~Millozzi, ``Large scale properties of
  the webgraph,'' \emph{The European Physical Journal B}, vol.~38, no.~2, pp.
  239--243, 2004.

\bibitem{gonzalezPowerGraphDistributedGraphParallel2012}
J.~E. Gonzalez, Y.~Low, H.~Gu, D.~Bickson, and C.~Guestrin,
  ``$\{$PowerGraph$\}$: Distributed $\{$Graph-Parallel$\}$ computation on
  natural graphs,'' in \emph{10th USENIX symposium on operating systems design
  and implementation (OSDI 12)}, 2012, pp. 17--30.

\bibitem{wanPipeGCNEfficientFullGraph2021}
C.~Wan, Y.~Li, C.~R. Wolfe, A.~Kyrillidis, N.~S. Kim, and Y.~Lin, ``Pipegcn:
  Efficient full-graph training of graph convolutional networks with pipelined
  feature communication,'' 2022.

\bibitem{ImprovingAccuracyScalability}
Z.~Jia, S.~Lin, M.~Gao, M.~Zaharia, and A.~Aiken, ``Improving the accuracy,
  scalability, and performance of graph neural networks with roc,''
  \emph{Proceedings of Machine Learning and Systems}, vol.~2, pp. 187--198,
  2020.

\bibitem{wangHeterogeneousGraphAttention2019}
X.~Wang, H.~Ji, C.~Shi, B.~Wang, P.~Cui, P.~Yu, and Y.~Ye, ``Heterogeneous
  graph attention network,'' 2019.

\bibitem{fuMAGNNMetapathAggregated2020}
X.~Fu, J.~Zhang, Z.~Meng, and I.~King, ``Magnn: Metapath aggregated graph
  neural network for heterogeneous graph embedding,'' \emph{arXiv preprint
  arXiv:2002.01680}, 2020.

\bibitem{huOpenGraphBenchmark2021}
W.~Hu, M.~Fey, M.~Zitnik, Y.~Dong, H.~Ren, B.~Liu, M.~Catasta, and J.~Leskovec,
  ``Open graph benchmark: Datasets for machine learning on graphs,''
  \emph{Advances in neural information processing systems}, vol.~33, pp.
  22\,118--22\,133, 2020.

\bibitem{fan2020incrementalization}
W.~Fan, M.~Liu, C.~Tian, R.~Xu, and J.~Zhou, ``Incrementalization of graph
  partitioning algorithms,'' \emph{Proceedings of the VLDB Endowment}, vol.~13,
  no.~8, pp. 1261--1274, 2020.

\bibitem{andreevBalancedGraphPartitioning2004}
K.~Andreev and H.~Räcke, ``Balanced graph partitioning,'' \emph{ACM}, 2004.

\bibitem{slotaPuLPScalableMultiobjective2015}
G.~M. Slota, K.~Madduri, and S.~Rajamanickam, ``Pulp: Scalable multi-objective
  multi-constraint partitioning for small-world networks,'' in \emph{IEEE
  International Conference on Big Data}, 2015.

\bibitem{cuthillReducingBandwidthSparse1969}
E.~Cuthill, ``Reducing the bandwidth of sparse symmetric matrices,'' in
  \emph{Acm National Conference}, 1969.

\bibitem{blandfordExperimentalAnalysisCompact}
D.~K. Blandford, G.~E. Blelloch, and I.~A. Kash, ``An experimental analysis of
  a compact graph representation,'' \emph{Dissertation Abstracts
  International}, 2004.

\bibitem{dhulipalaCompressingGraphsIndexes2016}
L.~Dhulipala, I.~Kabiljo, B.~Karrer, G.~Ottaviano, S.~Pupyrev, and A.~Shalita,
  ``Compressing graphs and indexes with recursive graph bisection,''
  \emph{ACM}, 2016.

\bibitem{SpeculativeParallelReverse}
D.~Mlakar, M.~Winter, M.~Parger, and M.~Steinberger, ``Speculative parallel
  reverse cuthill-mckee reordering on multi- and many-core architectures,'' in
  \emph{2021 IEEE International Parallel and Distributed Processing Symposium
  (IPDPS)}, 2021.

\bibitem{haoSpeedupGraphProcessing2016}
W.~Hao, J.~X. Yu, C.~Lu, and X.~Lin, ``Speedup graph processing by graph
  ordering,'' in \emph{the 2016 International Conference}, 2016.

\bibitem{araiRabbitOrderJustinTime2016}
J.~Arai, H.~Shiokawa, T.~Yamamuro, M.~Onizuka, and S.~Iwamura, ``Rabbit order:
  Just-in-time parallel reordering for fast graph analysis,'' in \emph{2016
  IEEE International Parallel and Distributed Processing Symposium (IPDPS)},
  2016.

\bibitem{balajiWhenGraphReordering2018}
V.~Balaji and B.~Lucia, ``When is graph reordering an optimization? studying
  the effect of lightweight graph reordering across applications and input
  graphs,'' in \emph{2018 IEEE International Symposium on Workload
  Characterization (IISWC)}, 2018.

\bibitem{barikVertexReorderingRealWorld2020}
R.~Barik, M.~Minutoli, M.~Halappanavar, N.~R. Tallent, and A.~Kalyanaraman,
  ``Vertex reordering for real-world graphs and applications: An empirical
  evaluation,'' in \emph{2020 IEEE International Symposium on Workload
  Characterization (IISWC)}, 2020.

\bibitem{boldiWebgraphFrameworkCompression2004}
P.~Boldi and S.~Vigna, ``The webgraph framework i: Compression techniques,'' in
  \emph{International World Wide Web Conference}, 2004.

\bibitem{vitterEfficientAlgorithmSequential1987b}
J.~S. Vitter, ``An efficient algorithm for sequential random sampling,''
  \emph{ACM transactions on mathematical software (TOMS)}, vol.~13, no.~1, pp.
  58--67, 1987.

\bibitem{walkerNewFastMethod1974}
A.~J. Walker, ``New fast method for generating discrete random numbers with
  arbitrary frequency distributions,'' \emph{Electronics Letters}, vol.~8,
  no.~10, pp. 127--128, 1974.

\bibitem{efraimidisWeightedRandomSampling2006}
P.~S. Efraimidis and P.~G. Spirakis, ``Weighted random sampling with a
  reservoir,'' \emph{Information processing letters}, vol.~97, no.~5, pp.
  181--185, 2006.

\bibitem{zhangAGLScalableSystem2020}
D.~Zhang, X.~Huang, Z.~Liu, J.~Zhou, Z.~Hu, X.~Song, Z.~Ge, L.~Wang, Z.~Zhang,
  and Y.~Qi, ``Agl: A scalable system for industrial-purpose graph machine
  learning,'' \emph{Proceedings of the VLDB Endowment}, vol.~13, no.~12.

\bibitem{yinDGIEasyEfficient2023}
P.~Yin, X.~Yan, J.~Zhou, Q.~Fu, Z.~Cai, J.~Cheng, B.~Tang, and M.~Wang, ``Dgi:
  An easy and efficient framework for gnn model evaluation,'' in
  \emph{Proceedings of the 29th ACM SIGKDD Conference on Knowledge Discovery
  and Data Mining}, 2023, pp. 5439--5450.

\bibitem{ZarrPythonZarr14}
\BIBentryALTinterwordspacing
Zarr-{{Python}} — zarr 2.14.2 documentation. [Online]. Available:
  \url{https://zarr.readthedocs.io/en/stable/}
\BIBentrySTDinterwordspacing

\bibitem{BloscMainBlog}
\BIBentryALTinterwordspacing
Blosc main blog page. [Online]. Available: \url{https://www.blosc.org/}
\BIBentrySTDinterwordspacing

\bibitem{linPaGraphScalingGNN2020}
Z.~Lin, C.~Li, Y.~Miao, Y.~Liu, and Y.~Xu, ``Pagraph: Scaling gnn training on
  large graphs via computation-aware caching,'' in \emph{SoCC '20: ACM
  Symposium on Cloud Computing}, 2020.

\bibitem{huOGBLSCLargeScaleChallenge2021}
W.~Hu, M.~Fey, H.~Ren, M.~Nakata, Y.~Dong, and J.~Leskovec, ``Ogb-lsc: A
  large-scale challenge for machine learning on graphs,'' in \emph{Thirty-fifth
  Conference on Neural Information Processing Systems Datasets and Benchmarks
  Track (Round 2)}, 2021.

\bibitem{Twitter2010}
\BIBentryALTinterwordspacing
Twitter-2010. [Online]. Available:
  \url{https://law.di.unimi.it/webdata/twitter-2010/}
\BIBentrySTDinterwordspacing

\bibitem{huHeterogeneousGraphTransformer2020}
Z.~Hu, Y.~Dong, K.~Wang, and Y.~Sun, ``Heterogeneous graph transformer,'' 2020.

\bibitem{tripathyReducingCommunicationGraph2020}
A.~Tripathy, K.~Yelick, and A.~Bulu{\c{c}}, ``Reducing communication in graph
  neural network training,'' in \emph{SC20: International Conference for High
  Performance Computing, Networking, Storage and Analysis}.\hskip 1em plus
  0.5em minus 0.4em\relax IEEE, 2020, pp. 1--14.

\bibitem{thuvanGDLLScalableShare2022}
D.~T.~T. Van, M.~N. Khan, T.~H. Afridi, I.~Ullah, A.~Alam, and Y.-K. Lee,
  ``Gdll: A scalable and share nothing architecture based distributed graph
  neural networks framework,'' \emph{IEEE Access}, vol.~10, pp.
  21\,684--21\,700, 2022.

\end{thebibliography}
\bibliographystyle{IEEEtran}

\end{document}